\documentclass[10pt,journal,compsoc]{IEEEtran}

\ifCLASSOPTIONcompsoc
  \usepackage[nocompress]{cite}
\else
  \usepackage{cite}
\fi

\ifCLASSINFOpdf
\else
\fi

\usepackage[hyphens]{url}

\usepackage{multirow}
\usepackage{graphicx}
\usepackage{amsmath,amsfonts}
\usepackage{caption}
\usepackage{cuted}
\usepackage[dvipsnames]{xcolor}
\usepackage{ragged2e}  %
\usepackage{xspace}
\usepackage{subcaption}  %
\usepackage{amsmath}
\usepackage{amssymb}
\usepackage{academicons}

\usepackage{acronym}
\usepackage{scalerel}
\usepackage{tikz}

\usepackage{tcolorbox}
\usepackage{clipboard}

\usepackage{balance}

\usepackage{hyperref} %
\usepackage[nameinlink]{cleveref} %

\usetikzlibrary{svg.path}

\definecolor{orcidlogocol}{HTML}{A6CE39}
\tikzset{
  orcidlogo/.pic={
    \fill[orcidlogocol] svg{M256,128c0,70.7-57.3,128-128,128C57.3,256,0,198.7,0,128C0,57.3,57.3,0,128,0C198.7,0,256,57.3,256,128z};
    \fill[white] svg{M86.3,186.2H70.9V79.1h15.4v48.4V186.2z}
                 svg{M108.9,79.1h41.6c39.6,0,57,28.3,57,53.6c0,27.5-21.5,53.6-56.8,53.6h-41.8V79.1z M124.3,172.4h24.5c34.9,0,42.9-26.5,42.9-39.7c0-21.5-13.7-39.7-43.7-39.7h-23.7V172.4z}
                 svg{M88.7,56.8c0,5.5-4.5,10.1-10.1,10.1c-5.6,0-10.1-4.6-10.1-10.1c0-5.6,4.5-10.1,10.1-10.1C84.2,46.7,88.7,51.3,88.7,56.8z};
  }
}

\newcommand\orcidicon[1]{\href{https://orcid.org/#1}{\mbox{\scalerel*{
\begin{tikzpicture}[yscale=-1,transform shape]
\pic{orcidlogo};
\end{tikzpicture}
}{|}}}}

\definecolor{koborderbright}{HTML}{d69509}
\definecolor{kobackbright}{HTML}{e8ddc5}
\definecolor{koborderteal}{HTML}{368be0}
\definecolor{kobackteal}{HTML}{dfe6e8}
\definecolor{repoboxborder}{HTML}{32a852}
\definecolor{repoboxback}{HTML}{bed1c3}
\newcommand{\gboxbegin}[4] {
    \begin{tcolorbox}[colback=#1,colframe=#2,title=\textbf{\emph{KEY #3 {#4}}}]
}
\newcommand{\gboxend} {
 \end{tcolorbox}
}

\newcounter{rtaskobsno}
\DeclareRobustCommand{\rtaskobs}[1]{%
   \refstepcounter{rtaskobsno}%
   \thertaskobsno\label{#1}}

\newcounter{rtasktakno}
\DeclareRobustCommand{\rtasktak}[1]{%
   \refstepcounter{rtasktakno}%
   \thertasktakno\label{#1}}

\newcommand{\mechanism}{{RedBit}\xspace}
\newcommand{\mechanismtitle}{{{Red\textcolor{red}{Bit}}\xspace}}

\newif\ifcameraready
\camerareadyfalse

\definecolor{observationblue}{rgb}{0.63, 0.87, 0.90}
\definecolor{aqua}{rgb}{0.0, 1.0, 1.0}

\ifcameraready

\else
    
    \definecolor{dblue}{rgb}{0.00, 0.00, 0.75}
    \definecolor{dgreen}{rgb}{0.00, 0.75, 0.00}

\fi

\newacro{cnn} [CNN] {Convolutional Neural Network}

\newacro{dnn} [DNN] {Deep Neural Network}

\newacro{qcnn} [QCNN] {Quantized Convolutional Neural Network}

\newacro{gpu} [GPU] {Graphics Processing Unit}

\newacro{pim} [PIM] {Processing-in-Memory}

\newacro{fpga} [FPGA] {Field-Programmable Gate Array}

\newacro{qnn} [QNN] {Quantized Neural Network}

\newacro{bcnn} [BCNN] {Binary Convolutional Neural Network}

\newacro{bnn} [BNN] {Binary Neural Network}

\newacro{bwn} [BWN] {Binary Weight Network}

\newacro{twn} [TWN] {Ternary Weight Network}

\newacro{ttq} [TTQ] {Trained Ternary Quantization}

\newacro{lca} [LCA] {Laboratory for Advanced Computing}

\newacro{ilsvrc} [ILSVRC] {ImageNet Large Scale Visual Recognition Challenge}

\newacro{asic} [ASIC] {Application Specific Integrated Circuit}

\newacro{soc} [SoC] {System-on-Chip}

\newacro{gemm} [GEMM] {General Matrix Multiplication}

\newacro{lut} [LUT] {Lookup Table}

\newacro{hpc} [HPC] {high-performance computing}

\newacro{cpu} [CPU] {Central Processing Unit}

\begin{document}

\title{\mechanismtitle: An End-to-End Flexible Framework \\ for Evaluating the Accuracy of Quantized CNNs}

\newcommand{\orcid}[1]{\href{https://orcid.org/#1}{\textcolor[HTML]{A6CE39}{\aiOrcid}}}

\author{André~Santos$^{\orcidicon{}}$,~\IEEEmembership{Student Member,~IEEE,} \and
        João~Dinis~Ferreira$^{\orcidicon{0000-0003-3855-3906}}$,~\IEEEmembership{Student Member,~IEEE,} \and
        Onur~Mutlu$^{\orcidicon{0000-0002-0075-2312}}$,~\IEEEmembership{Fellow,~IEEE,} \and
        and~Gabriel~Falcao$^{\orcidicon{0000-0001-9805-6747}}$,~\IEEEmembership{Senior Member,~IEEE}%

\IEEEcompsocitemizethanks{
\IEEEcompsocthanksitem A. Santos and G. Falcao are with the University of Coimbra and Instituto de Telecomunicações, at the Department of Electrical and Computer Engineering, Coimbra, Portugal.

J.D. Ferreira and O. Mutlu are with the Department of Electrical Engineering and Information Technology of ETH Zürich, Switzerland, and the SAFARI Research Group.
\protect\\}%
}

\IEEEtitleabstractindextext{%

\begin{abstract}

\justifying{In recent years, \acfp{cnn} have become the standard class of deep neural network for image processing, classification and segmentation tasks.
However, the large strides in accuracy obtained by \acp{cnn} have been derived from increasing the complexity of network topologies, which incurs sizeable performance and energy penalties in the training and inference of \acp{cnn}.
Many recent works have validated the effectiveness of \emph{parameter quantization}, which consists in reducing the bit width of the network's parameters, to enable the attainment of considerable performance and energy efficiency gains without significantly compromising accuracy.}

\justifying
\setlength{\parindent}{0pt}
\setlength{\parskip}{1pt}

However, it is difficult to compare the relative effectiveness of different quantization methods.
To address this problem, we introduce \mechanism, an open-source framework that provides a transparent, extensible and easy-to-use interface to evaluate the effectiveness of different algorithms and parameters configurations on network accuracy.

We use \mechanism to perform a comprehensive survey of five state-of-the-art quantization methods applied to the MNIST, CIFAR-10 and ImageNet datasets. We evaluate a total of $2300$ individual bit width combinations, independently tuning the width of the network's weight and input activation parameters, from $32$ bits down to $1$ bit (e.g., 8/8, 2/2, 1/32, 1/1, for weights/activations).
Upwards of 20000 hours of compute time in a pool of state-of-the-art GPUs were used to generate all the results in this paper.
For 1-bit quantization, the accuracy losses for the MNIST, CIFAR-10 and ImageNet datasets range between $[0.26\%, 0.79\%]$, $[9.74\%, 32.96\%]$ and $[10.86\%, 47.36\%]$ top-1, respectively.
We actively encourage the reader to download the source code and experiment with \mechanism, and to submit their own observed results to our public repository, available at \href{https://github.com/IT-Coimbra/RedBit}{https://github.com/IT-Coimbra/RedBit}.

\end{abstract}

}

\maketitle

\IEEEdisplaynontitleabstractindextext

\IEEEpeerreviewmaketitle

\IEEEraisesectionheading{\section{Introduction}\label{sec:introduction}}

\IEEEPARstart{I}{n} recent years, \acp{cnn} have become increasingly adept at executing numerous complex image processing, classification and segmentation tasks.
These improvements have been attained in large part at the expense of a continuous increase in size and complexity for new \ac{cnn} network topologies.
ResNet-50~\cite{resnet} (introduced in 2015, with 26 million parameters), achieves a top-1 accuracy of 77.15\%, and a top-5 accuracy of 93.29\%.
(Top-N accuracy corresponds to the proportion of scenarios for which the correct answer is contained in the network's $N$ best guesses, for a given classification problem.)
In contrast, EfficientNet-B7~\cite{efficientnet} (introduced in 2019, with 66 million parameters, $2.5 \times$ larger than ResNet-50), achieves top-1 and top-5 accuracies for the ImageNet dataset of 84.40\% and 97.10\%, respectively.
The upshot of this increase in the computational complexity is a substantial increase in the time and energy required to train and use them.

The widespread adoption of \acp{cnn} is potentially most impactful in edge devices (e.g., autonomous vehicles, smartphones), which often come equipped with high-quality imaging sensors.
However, these devices also carry very strict autonomy constraints, and are therefore unsuited to execute the complex and memory-intensive operations associated with conventional \acp{cnn}.
The severity of this issue will continue to escalate in the near future, as newly proposed networks make use of increasingly large numbers of parameters, exacerbating their memory intensity and performance and energy overheads~\cite{energy-problem}.

To curb the scaling challenges \cite{challenges} presented by these larger networks, while retaining as many of their benefits as possible, it is possible to \emph{quantize} \ac{cnn} parameters, yielding \acfp{qcnn}.
Quantization consists in reducing the bit width of a network's parameters to alleviate their computational and data movement requirements, and has been demonstrated by many prior works \cite{qnn,dorefa-net,xnor-net,twn,ttq,vector_quantization,qcnn_for_mobile,incremental_quantization,hwgq_net,lq_nets,bi_real,tow_step_quantization,multiple_step_quantization,bwnh,pact,wrpn,syq,dist_quant,apprentice,loss_aware_binarization,defensive_quantization,reg_training_bnn,circulant_bcnns,bnn_plus,blended_coarse_gradient,proxquant,self_binarizing_net,acc_compact_bnn,bbg,bnn_r_to_b_conv,info_retention,pami_wang_learning_2021,pami_wang_gradient_2021,pami_bulat_hierarchical_2020,pami_sun_lazily_2020,pami_zhuang_effective_2021,pami_tang_towards_2020,pami_han_learning_2021,pami_li_learning_2021,pami_duan_learning_2019} to provide substantial performance and energy gains, with minimal losses in accuracy.
Quantization allows 1) the use of less compute resources - less bits involved in logic operations - 2) the use of less memory to store the actual data and 3) the reduction of data movement. These lead to reductions on area footprints for computation hardware implementation and required memory.

Other techniques can also alleviate computational requirements such as pruning \cite{brain_damage, automatic_pruning, neuron_pruning, pruning, channel_pruning}, fine-tuning \cite{fine_tuning}, compression \cite{deep_compression, prune_bin, deep_compression2}, decomposition \cite{low_rank_decomposition, circulant_projections, cp_decomposition}, knowledge distillation/transfer learning \cite{knowledge_distillation, kd_and_tl, abc_net, kd_adversarial_networks, darkrank}, or others \cite{cpu_optimization, linear_structure}, but this work only focus on quantization.

Many recent works have demonstrated how it is possible to leverage \acp{qcnn} to greatly accelerate the training and inference in many specialized architectures, including \acp{gpu}~\cite{tensor-core, mixed_prec_training, mixed_prec_training2, mixed_prec_solvers, bfloat16, ampt_ga, mp_solver_matrix, mp_cluster, mp_cluster2, mp_cluster3}, \acp{fpga}~\cite{quantizationFPGA, openCL-FPGA-accelerator, scnn-fpga-accelerator, mnist-fpga-accelerator, dsc-fpga-accelerator}, \acp{asic}~\cite{yodann-asic-accelerator, cnn-fpga-asic-accelerator, deepopt-asic, dnn_accelerator} and low-power \acp{soc}~\cite{asic_soc}.

A particular case of interest is binarization, i.e., the quantization of all parameters to 1-bit~\cite{review-bnns, bnn-survey}.
In \acp{bcnn}, convolution operations can be performed exclusively with resort to bitwise logic operations, which enables high performance improvements~\cite{BNN_FPGA_implementation1,BNN_FPGA_implementation2}.
This has been especially useful for emerging computing paradigms (e.g., \ac{pim}) which achieve very high throughput and energy efficiency for bitwise operations~\cite{drisa,ambit,neural-cache,dima,rebnn,rtl_lib-compiler}.

It is vital to understand the effect of quantizing \ac{cnn} parameters, and the benefits and drawbacks of each of the quantization methods proposed in prior works.
In this work, we not only present quantization methods and their results, but also study and compare them, by 1) fully training them with different weight/activation quantization levels, and 2) evaluating their performance for the MNIST, CIFAR-10 and ImageNet datasets.

\textbf{To this end}, we introduce \textbf{\textit{\mechanism}}, a PyTorch-based, open-source framework that enables the design space exploration of different quantization parameters and their impact on network accuracy.
Using \mechanism, we evaluate the accuracy of five leading quantization methods, as determined by a thorough survey of the literature, across $4500$ intermediate training processes.
Through this analysis, we identify the quantization hyperparameters that maximize accuracy.
Our results were generated over upwards of $20000$ hours of compute time on a pool of state-of-the-art GPUs, resulting in the training of over $2300$ unique network models using different quantization methods and parameters.

All our results are available in the public Git repository, \href{https://github.com/IT-Coimbra/RedBit}{https://github.com/IT-Coimbra/RedBit}.
We highly encourage readers to download and use the provided source code, and also to contribute their own GPU time to run experiments using \mechanism, and to upload their accuracy results to our repository.
Through these contributions, it is possible to make progress towards our end goal of quantifying the accuracy provided by as many quantization methods as possible.
We also encourage readers to contribute to the development of the framework with improvements, e.g., support for more algorithms, quantization methods and further optimizations.

\section{Background}
\label{sec:background}

This section provides an overview of key concepts related to state-of-the-art \ac{cnn} architectures and commonly used datasets, and introduces the concept of quantization and the quantization algorithms which we later analyze.

\subsection{Convolutional Neural Networks}
\label{sec:convolutional-neural-networks}

\acp{cnn} are a class of deep neural networks often applied image-related machine learning tasks.
Their name is derived from the fact that most of the layers in \acp{cnn} are \emph{convolutional layers}, wherein the convolution operation is performed between a given input and a (smaller) feature detector.
Batch normalization, non-linearity and pooling layers are also present in \acp{cnn}. AlexNet~\cite{alexnet}, ResNet~\cite{resnet}, DenseNet~\cite{densenet}, VGG \cite{vgg}, Inception class \cite{inception}, Deep Networks \cite{deep_network}, MobileNet \cite{mobilenet_v1, mobilenet_v2}, ShuffleNet \cite{shufflenet_v1, shufflenet_v2}, SqueezeNet \cite{squeezenet} and EfficientNet~\cite{efficientnet} are popular types of \acp{cnn}.

\acp{cnn} can train successfully because of the back propagation algorithm \cite{back_propagation1, back_propagation2, back_propagation3, back_propagation4}.

\begin{figure*}[!htb]
\centering
\begin{tabular}{ccc}
\subcaptionbox{\label{subfig:1D-conv-operation}}{\includegraphics[height = 30mm]{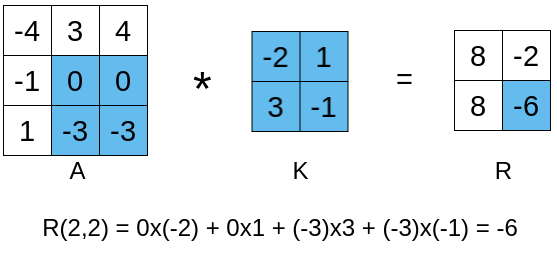}} &
\subcaptionbox{\label{subfig:convolution-kernel-movement}}{\includegraphics[height = 35mm]{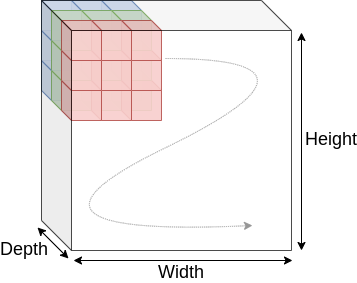}} &
\subcaptionbox{\label{subfig:shortcut}}{\includegraphics[height = 35mm]{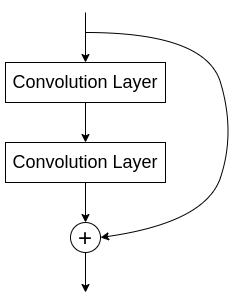}}
\end{tabular}
\caption{\textbf{(a)} 1D convolution operation. Each element from input A is multiplied to the corresponding element in the kernel K, i.e., point-wise multiplication. The final result value is the sum of all multiplications. \textbf{(b)} Kernel movement in a convolutional layer. Depth represents the number of input channels. \textbf{(c)} Basic block in the ResNet CNN architecture.
}
\end{figure*}

\subsubsection{The Convolutional/Fully connected Layers}
\label{sec:conv-fully-connected-layers}

In a convolutional layer the convolution operation is applied between a 3D kernel and the inputs of that layer.
The convolution operation is a point-wise product followed by an addition reduction as it is shown in Figure \ref{subfig:1D-conv-operation}.

CNNs also have fully connected layers, where the convolution operation also takes place.
Throughout this work, the kernel parameters values are referred to as weights and the inputs of a layer referred to as input activations. Also, when we mention input activations, we are generally referring to the inputs of convolutional/fully connected layers.

On \acp{cnn}, the convolution operation is expanded into 3D, as depicted in Figure \ref{subfig:convolution-kernel-movement}. This figure also shows how the kernel moves throughout the input. In this example, the input is an RGB image.

\subsubsection{The Non-linearity Layer}

With only convolutional/fully connected layers, it is difficult for a neural network to approach the solution for a problem.
With the use of non-linear layers, it is proven that a neural network can be fitted to solve non-linear problems~\cite{non-linearity-cnn}.
Examples of popular non-linear functions used in CNNs are Sigmoid, ReLU and HardTanh.

\subsubsection{The Batch Normalization Layer}

To train \acp{cnn} more effectively, Ioffe and Szegedy~\cite{batchnorm} introduce a type of layer known as batch normalization.
Batch normalizations re-centers and re-scales input activations.
The normalization helps reduce the internal covariate shift present in the input activations during training allowing the model to converge faster to the optimal solution, by using bigger values for the learning rate parameter.

\subsubsection{The Pooling Layers}
\label{sec:pooling-layers}

Along the neural network, the number of feature maps, i.e., number of output channels of a layer, often changes and also their size.
This change is often referred to as downsampling and help reduce the computational requirements.
To downsample, CNNs make use of pooling layers.
Most popular types are average pooling and maximum pooling, where a kernel passing across a feature map pools the average or maximum value, respectively.

The output of pooling layers are also summarized versions of the input feature maps, making them more robust.

\subsubsection{Shortcuts}

Some \acp{cnn} (e.g., ResNet) employ \emph{shortcuts}, which facilitate the flow of information across the network.
The key idea is to pass the input activations not only to the next layer, but also to the subsequent layers, as shown in \Cref{subfig:shortcut}.

\subsection{Datasets}
\label{sec:datasets}

We evaluate the performance of several \acp{cnn} for three popular datasets: MNIST, CIFAR-10 and ImageNet. This section succinctly describes the characteristics of each of these datasets.

\vspace{1mm}
\noindent
\textbf{MNIST}~\cite{lenet-5} contains 70,000 28x28 grey images of handwritten digits. 60,000 images compose the training set and the remain 10,000 belong to the test set.
The digits have been size normalized and centered in a fixed-size image.

\vspace{1mm}
\noindent
\textbf{CIFAR-10}~\cite{cifar-10} contains 60,000 32x32 color images equally distributed across the following 10 classes: \{airplane, automobile, bird, cat, deer, dog, frog, horse, ship, truck\}.
This dataset contains 50,000 images in the training set and 10,000 images in the test set.

\vspace{1mm}
\noindent
\textbf{ImageNet}~\cite{imagenet} contains over 14 million images of various sizes, across 1000 classes.
There are multiple revisions of this dataset. In this work we use the \ac{ilsvrc} 2012 version.

\section{Leading Quantization Methods}

\begin{figure*}[!hbt]
	\centering
	\includegraphics[width =\textwidth]{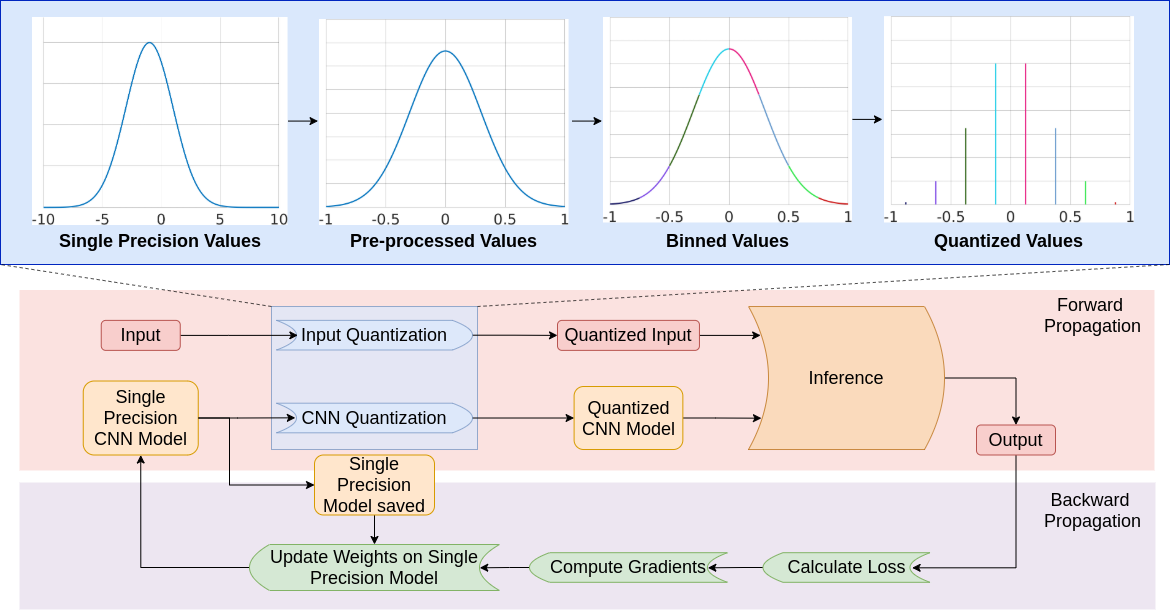}
	\caption{Quantization process and training cycle.}
	\label{fig:quant-process}
\end{figure*}

This section presents 5 quantization methods that are studied in more depth in the following sections. This section intends to present how quantization of \acp{cnn} happen and key characteristics that describe each quantization method. Figure \ref{fig:quant-process} shows how the quantization process occurs, in general. 

Typical \acp{cnn} use 32-bit floating point parameters. Quantization can reduce by up to 32x the model size.
Binarization, is a particular case of quantization that has been widely studied~\cite{qnn,xnor-net,dorefa-net,bi_real,reg_training_bnn,circulant_bcnns,bnn_plus,self_binarizing_net,acc_compact_bnn}.
When both weights and input activations are binarized, convolution operations can be simplified to bit-wise logic operations making them more computationally efficient.
These optimizations can be leveraged to develop efficient specialized accelerators (e.g., \ac{fpga}-~\cite{rtl_lib-compiler,quantizationFPGA,openCL-FPGA-accelerator, BNN_FPGA_implementation1, BNN_FPGA_implementation2}, \ac{asic}-~\cite{yodann-asic-accelerator, cnn-fpga-asic-accelerator, deepopt-asic, dnn_accelerator}, and \ac{pim}-based~\cite{rowclone,bitwise-operations,pim-instructions,drisa,ambit,neural-cache,dima,parapim,rebnn,ferreira2021pluto}).

Due to the discretization of weights and/or input activations, the back propagation algorithm needs to be adjusted to properly determine the gradients of discrete functions. The straight-through estimator \cite{ste1, ste2} is mostly used by the following quantization methods to successfully obtain the correct gradients to update the single precision weights.

\subsection{QNN: Quantized Neural Networks} %

Courbariaux et al.~\cite{lpm,binary_connect,binarized_nn} study the impact of quantization in the multiplications that occur in \aclp{dnn} by employing three different representations during the training phase: floating point, fixed point, and dynamic fixed point.
Building upon these studies, Hubara et al.~\cite{qnn} present a method to train \acp{qnn}.
In this work, the authors study the effect of binarizing both the network's weights and input activations, and demonstrate that the quantization of gradients results in minimal accuracy degradation.
The main focus of this work is the binarization of weights and input activations, but it also studies other levels of quantization using the linear method described in equation \ref{eq:linear-quant} where $minV$ and $maxV$ are the minimum and maximum scale range, respectively.
\begin{equation} \label{eq:linear-quant}
\begin{split}
	LinearQuant \left.(x,bit width\right) = \\
	Clip \left(\frac{round \left(x \times 2^{bit width-1}\right)}{2^{bit width-1}}, minV, maxV \right) 
\end{split}
\end{equation}

This was the first work to successfully quantize all layers in CNNs without significant accuracy degradation.
It achieves 99.04\% accuracy on the MNIST validation set and 88.60\% accuracy on the CIFAR-10 validation set while applying binarization to both weights and input activations. In the ImageNet validation set, using AlexNet, it achieves 56.6\% top-1 and 80.2\% top-5 accuracy, in single-precision.
Binarizing both weights and input activations reduces the accuracy to 41.8\% top-1 and 67.1\% top-5. With 2-bit quantization of activations, keeping weights binary, the accuracy increases to 51.03\% top-1 and 73.67\% top-5.

Both weights and input activations are binarized using the function

\begin{equation} \label{eq:sign}
	q=\operatorname{sign}(r)=\left\{\begin{array}{ll}
		+1 & \text { if } r \geq 0 \\
		-1 & \text { otherwise }
	\end{array}\right.,
\end{equation}

\noindent
where $r$ is a single precision value and $q$ can be represented as a 1-bit value, i.e., binary value.

\subsection{XNOR-Net: Scaling Factor}

Rastegari et al.~\cite{xnor-net} show that in order to minimize the Euclidian distance between single-precision weights and binarized weights, the tensor of single-precision weights $W$ can be approximated via the product $W \approx \alpha B$, where $\alpha$ is a scaling factor, and $B$ is a tensor of binary weights.
Here, the scaling factors are single-precision values that serve to expand the range of representability when multiplied with normalized values.
Using this scaling-factor-based approximation, the authors introduce the \ac{bwn}, whose weights are binarized, and the XNOR-Net, whose weights and input activations are binarized.
This work further demonstrates that an optimal scaling factor can be given by \Cref{eq:optimal-alpha}, where $W$ is a vector in $\mathbb{R}^n$, $n=c \times w \times h$. $W$ contains the single-precision weight values.
An optimal estimation of a binary weight can be simply achieved applying equation \ref{eq:sign}~\cite{qnn}.

\begin{equation} \label{eq:optimal-alpha}
	\alpha^{*}=\frac{\mathbf{W}^{\top} \operatorname{sign}(\mathbf{W})}{n}=\frac{\sum\left|\mathbf{W}_{i}\right|}{n}=\frac{1}{n}\|\mathbf{W}\|_{\ell 1}
\end{equation}

\cite{xnor-net} also shows that scaling factors associated with input activations can improve accuracy, but by less than 1\%.
For this reason, the authors do not use it. Likewise, we also do not implement it in our framework; instead, we only use scaling factors in association with the network's weights.

In single-precision, AlexNet achieves 56.6\% top-1 and 80.2\% top-5 accuracy on the ImageNet validation set, according to the authors.
When applying \ac{bwn}, i.e., binarizing only weights and using scaling factors, there is no accuracy loss, with final accuracy values of 56.8\% for top-1 and 79.4\% for top-5.
If XNOR-Net is applied, input activations are also binarized, and the accuracy decreases to 44.2\% for top-1 and 69.2\% for top-5.

\subsection{DoReFa-Net: Low Bit Width Parameters}

Zhou et al.~\cite{dorefa-net} took some ideas from the works of Hubara et al.~\cite{qnn} and Rastegari et al.~\cite{xnor-net} and propose a new method, DoReFa-Net, to quantize weights, activations and gradients.

Similar to the work of Rastegari et al. in~\cite{xnor-net}, the first and last layers are not quantized in DoReFa-Net.
When binarizing weights, DoReFa-Net employs use a simpler approach than XNOR-Net, making it so that each scaling factor is equal to the mean of absolute values of each output channel of weights, instead of applying \Cref{eq:optimal-alpha}.

This work also confirms that, while quantizing gradients, they need to be stochastically quantized instead of deterministically, as it was pointed out by Hubara et al. in~\cite{qnn}.

According to its authors, DoReFa-Net applied to AlexNet achieves 55.9\% top-1 accuracy in single-precision.
When applying 8-bit quantization, it achieves 53.0\% top-1. It also achieves 40.1\% top-1, 47.7\% top-1 and 50.3\% top-1 accuracy when binarizing the weights and quantizing the activations to 1, 2 or 4 bits, respectively.

\subsection{Ternary Quantization: TWN and TTQ}

Hubara et al.~\cite{qnn} and Zhou et al.~\cite{dorefa-net} show that quantizing weights or activations to larger values than 1-bit can regain some accuracy lost during quantization, compared to binarization, approaching single-precision results.

Li et al.~\cite{twn} present \acp{twn} that apply ternarization to the weights of \acp{cnn}, i.e., converts 32-bit weights into -1, 0 and 1 (2-bit representation) values.

\ac{twn} only applies ternarization to the weights, keeping activations and gradients unchanged.
In order to improve the network even further, \ac{twn} seeks to minimize the Euclidian distance between the single-precision weights $W$ and the ternary-values weights $W^t$ along with a non-negative scaling factor $\alpha$, 
similar to the idea first introduced by Rastegari et al. in~\cite{xnor-net}.
The authors of \ac{twn} arrive at an approximated solution with a threshold-based ternary function (equation \ref{eq:2}). $\Delta$ is a positive threshold parameter.

\begin{equation} \label{eq:2}
	\mathrm{W}_{i}^{t}=f_{t}\left(\mathrm{W}_{i} | \Delta\right)=\left\{\begin{aligned}
		+1, & \text { if } \mathrm{W}_{i}>\Delta \\
		0, & \text { if }\left|\mathrm{W}_{i}\right| \leq \Delta \\
		-1, & \text { if } \mathrm{W}_{i}<-\Delta
	\end{aligned}\right.
\end{equation}

\ac{twn} applied to LeNet-5 achieves 99.35\% accuracy on MNIST validation set compared to 99.41\% achieved in single-precision. 
Applied to ResNet-18B, \ac{twn} achieves 65.3\% top-1 and 86.2\% top-5 accuracy on the ImageNet validation set, compared to 67.6\% top-1 and 88.0\% top-5 accuracy obtained in single-precision.

Zhu et al.~\cite{ttq} present \ac{ttq}, a similar concept as \ac{twn}.
\ac{ttq} also ternarizes weights to -1, 0 and 1 using a symmetric threshold.
Additionally, and where it starts to differ from \ac{twn}, it uses two scaling factors for positive and negative weights in each layer.
Also, these scaling factors are learned during training, instead of calculated.
Since the weights and scaling factors are trained, there are two independent gradients, one for weights and the other for scaling factors.

Applying \ac{ttq} to ResNet-20 and ResNet-56, it achieves 91.13\% and 93.56\% accuracy, respectively, on the CIFAR-10 validation set, compared to single-precision accuracy of 91.77\% and 93.2\%, respectively. On AlexNet, it achieves 57.5\% top-1 and 79.7\% top-5 on the ImageNet validation set, compared to 57.2\% top-1 and 80.3\% in single-precision.

\begin{table}[]
	\centering
	\begin{tabular}{|c|c|c|c|}
		\hline
		Method     & \begin{tabular}[c]{@{}c@{}}First and Last\\ Conv/FC\\ Layers Quantized\end{tabular} & \begin{tabular}[c]{@{}c@{}}Scaling \\ Factor(s)\end{tabular} & \begin{tabular}[c]{@{}c@{}}Layer \\ Reordering\end{tabular} \\ \hline
		QNN~\cite{qnn}               & \checkmark   & X               & X                \\ \hline
		DoReFa-Net~\cite{dorefa-net} & X            & \checkmark *     & X                \\ \hline
		XNOR-Net~\cite{xnor-net}     & X            & \checkmark      & \checkmark       \\ \hline
		TWN~\cite{twn}               & \checkmark   & \checkmark      & X                \\ \hline
		TTQ~\cite{ttq}               & X            & \checkmark      & X                \\ \hline
	\end{tabular}
	\caption{Key points used by each quantization method. Conv - Convolutional; FC - Fully Connected. * - Only when weights are binary}
	\label{tab:key-points}
\end{table}

\begin{table}[]
\centering
\renewcommand{\tabcolsep}{2pt}
\begin{tabular}{|c|c|c|ccc|c|}
\hline
\multirow{2}{*}{Neural Network} & \multirow{2}{*}{Work} & \multirow{2}{*}{Type} & \multicolumn{3}{c|}{Bit Width} & \multirow{2}{*}{Accuracy} \\
                                &                       &                       & W        & A        & G        &                                      \\ \hline
\multicolumn{7}{|c|}{Single-Precision}                                                                                                                  \\ \hline
LeNet-5                         &~\cite{qnn}            & Floating-Point        & 32       & 32       & 32       & 99.41\%                              \\ \hline
\multicolumn{7}{|c|}{Quantization}                                                                                                                      \\ \hline
LeNet-5                         &~\cite{twn}            & TWN                   & 2        & 32       & 32       & 99.35\%                              \\ \hline
\multicolumn{7}{|c|}{Binarization}                                                                                                                      \\ \hline
Multilayer Perceptron *1        &~\cite{qnn}            & BNN                   & 1        & 1        & 32       & 98.60\%                              \\
Multilayer Perceptron *2        &~\cite{qnn}            & BNN                   & 1        & 1        & 32       & 99.04\%                              \\ \hline
LeNet-5                         &~\cite{twn}            & BPWN                  & 1        & 32       & 32       & 99.05\%                              \\ \hline
\end{tabular}
\caption{State-of-the-art results obtained in multiple papers concerning the MNIST dataset. Despite the same Neural Network on various results, the Neural Network can have differences and/or the training method is different. *1: Neural Network implemented using the Torch7 framework. *2: Neural Network implemented using the Theano framework}
\label{tab:MNIST-t-results}
\end{table}

\begin{table}[]
\centering
\renewcommand{\tabcolsep}{2pt}
\begin{tabular}{|c|c|c|ccc|c|}
\hline
\multirow{2}{*}{Neural Network} & \multirow{2}{*}{Work} & \multirow{2}{*}{Type} & \multicolumn{3}{c|}{Bit Width} & \multirow{2}{*}{Accuracy} \\
                                &                       &                       & W        & A        & G        &                                      \\ \hline
\multicolumn{7}{|c|}{Single-Precision}                                                                                                                  \\ \hline
ResNet-20                       &~\cite{ttq}            & Floating-Point        & 32       & 32       & 32       & 91.77\%                              \\
ResNet-32                       &~\cite{ttq}            & Floating-Point        & 32       & 32       & 32       & 92.33\%                              \\
ResNet-44                       &~\cite{ttq}            & Floating-Point        & 32       & 32       & 32       & 92.82\%                              \\
ResNet-56                       &~\cite{ttq}            & Floating-Point        & 32       & 32       & 32       & 93.20\%                              \\ \hline
\multicolumn{7}{|c|}{Quantization}                                                                                                                      \\ \hline
ResNet-20                       &~\cite{ttq}            & TTQ                   & 2        & 32       & 32       & 91.13\%                              \\
ResNet-32                       &~\cite{ttq}            & TTQ                   & 2        & 32       & 32       & 92.37\%                              \\
ResNet-44                       &~\cite{ttq}            & TTQ                   & 2        & 32       & 32       & 92.98\%                              \\
ResNet-56                       &~\cite{ttq}            & TTQ                   & 2        & 32       & 32       & 93.56\%                              \\ \hline
\multicolumn{7}{|c|}{Binarization}                                                                                                                      \\ \hline
ConvNet                         &~\cite{qnn}            & BNN                   & 1        & 1        & 32       & 89.85\%                              \\
ConvNet                         &~\cite{qnn}            & BNN                   & 1        & 1        & 32       & 88.60\%                              \\
\hline
\end{tabular}
\caption{State-of-the-art results obtained in multiple papers concerning the CIFAR-10 dataset. }
\label{tab:CIFAR10-t-results}
\end{table}

\begin{table}[]
\centering
\renewcommand{\tabcolsep}{2pt}
\begin{tabular}{|c|c|c|ccc|cc|}
\hline
\multirow{2}{*}{Neural Network} & \multirow{2}{*}{Work} & \multirow{2}{*}{Type} & \multicolumn{3}{c|}{Bit Width} & \multicolumn{2}{c|}{Accuracy} \\
                                &                       &                       & W        & A        & G        & Top-1               & Top-5              \\ \hline
\multicolumn{8}{|c|}{Single-Precision}                                                                                                                      \\ \hline
AlexNet                         &~\cite{ttq}            & Floating-Point        & 32       & 32       & 32       & 57.2\%              & 80.3\%             \\
ResNet-18                       &~\cite{twn}            & Floating-Point        & 32       & 32       & 32       & 65.4\%              & 86.76\%            \\
ResNet-18B                      &~\cite{ttq}            & Floating-Point        & 32       & 32       & 32       & 69.6\%              & 89.2\%             \\
GoogLeNet                       &~\cite{qnn}            & Floating-Point        & 32       & 32       & 32       & 71.6\%              & 91.2\%             \\
\hline
\multicolumn{8}{|c|}{Quantization}                                                                                                                          \\ \hline
AlexNet                         &~\cite{dorefa-net}     & DoReFa-Net            & 8        & 8        & 8        & 53\%                & -                  \\
AlexNet                         &~\cite{ttq}            & TTQ                   & 2        & 32       & 32       & 57.5\%              & 79.7\%             \\
ResNet-18                       &~\cite{twn}            & TWN                   & 2        & 32       & 32       & 61.8\%              & 84.2\%             \\
ResNet-18B                      &~\cite{twn}            & TWN                   & 2        & 32       & 32       & 65.3\%              & 86.2\%             \\
ResNet-18B                      &~\cite{ttq}            & TTQ                   & 2        & 32       & 32       & 66.6\%              & 87.2\%             \\
GoogLeNet                       &~\cite{qnn}            & QNN                   & 4        & 4        & 32       & 66.5\%              & 83.4\%             \\
GoogLeNet                       &~\cite{qnn}            & QNN                   & 6        & 6        & 6        & 66.4\%              & 83.1\%             \\ \hline
\multicolumn{8}{|c|}{Binarization}                                                                                                                          \\ \hline
AlexNet                         &~\cite{xnor-net}       & BWN                   & 1        & 32       & 32       & 56.8\%              & 79.4\%             \\
AlexNet                         &~\cite{dorefa-net}     & DoReFa-Net            & 1        & 4        & 32       & 50.3\%              & -                  \\
AlexNet                         &~\cite{qnn}            & QNN                   & 1        & 2        & 32       & 51.03\%             & 73.67\%            \\
AlexNet                         &~\cite{dorefa-net}     & DoReFa-Net            & 1        & 2        & 32       & 47.7\%              & -                  \\
AlexNet                         &~\cite{qnn}            & BNN                   & 1        & 1        & 32       & 41.8\%              & 67.1\%             \\
AlexNet                         &~\cite{dorefa-net}     & DoReFa-Net            & 1        & 1        & 32       & 40.1\%              & -                  \\
AlexNet                         &~\cite{xnor-net}       & XNOR-Net              & 1        & 1        & 32       & 44.2\%              & 69.2\%             \\
ResNet-18B                      &~\cite{xnor-net}       & BWN                   & 1        & 32       & 32       & 60.8\%              & 83.0\%             \\
ResNet-18B                      &~\cite{xnor-net}       & XNOR-Net              & 1        & 1        & 32       & 51.2\%              & 73.2\%             \\
GoogLeNet                       &~\cite{qnn}            & BNN                   & 1        & 1        & 32       & 47.1\%              & 69.1\%             \\
GoogLeNet                       &~\cite{xnor-net}       & BWN                   & 1        & 32       & 32       & 65.5\%              & 86.1\%             \\ \hline
\end{tabular}
\caption{State-of-the-art results obtained in multiple papers concerning the ImageNet dataset.
}
\label{tab:ImageNet-t-results}
\end{table}

Table \ref{tab:key-points} shows major features implemented by previous works mentioned.
Tables \ref{tab:MNIST-t-results}, \ref{tab:CIFAR10-t-results} and \ref{tab:ImageNet-t-results} gather results obtained and previously mentioned by the works here depicted. Each method in their original work might select the same \acp{cnn} but, the implemented model structure in each work might have small differences between them. Due to these circumstances, it is difficult to fairly compare quantization results provided by each work and depicted on these tables.

\section{A Methodology for Systematically \\ Evaluating Quantization Methods}
\label{methodology}

This section simultaneously describes the \mechanism framework and demonstrates its application for the evaluation the following five state-of-the-art \ac{cnn} quantization methods: \ac{qnn}~\cite{qnn}, DoReFa-Net~\cite{dorefa-net},
XNOR-Net (including the \ac{bwn} variant)~\cite{xnor-net}, \ac{twn}~\cite{twn} and \ac{ttq}~\cite{ttq}.
The baseline \ac{cnn} is the same for all quantization methods.
We do not evaluate the effects of gradient quantization.

The source code of \mechanism is publicly available online, and we actively encourage the community to explore it and to contribute to its further development and testing.
The ultimate goal of \mechanism is to become a unified framework for reliably comparing Quantized Convolutional Neural Networks by aggregating results for many configurations. %
Interested readers may refer to the following repository:  \href{https://github.com/IT-Coimbra/RedBit}{https://github.com/IT-Coimbra/RedBit}.

\subsection{Implemented CNNs}

In this paper we show results only for a subset of networks for the CIFAR-10 and ImageNet datasets, due to the computational complexity required to train a greater number of models.
For the MNIST dataset, we evaluate LeNet-5.
For the CIFAR-10 dataset, we evaluate ResNet-20, ResNet-50, and VGG-16.
For the ImageNet dataset, we evaluate AlexNet, ResNet-18 and VGG-16.
We note, however, that \mechanism can be used to train any variant of these \acp{cnn}.
All associated implementation details can be found under the sections devoted to "CNN Architectures" in \href{https://github.com/IT-Coimbra/RedBit/wiki}{\mechanism's GitHub wiki.}

\subsection{Apparatus}
\label{apparatus}

\begin{table}[b]
\centering
\resizebox{0.9\columnwidth}{!}{
\renewcommand{\tabcolsep}{2pt}
\begin{tabular}{|c|c|c|c|c|}
\hline
\multicolumn{5}{|c|}{Hardware}                                                                                                                                                                                                                                        \\ \hline
Machine & \begin{tabular}[c]{@{}c@{}}CPU\\ (\#cores/threads)\end{tabular} & RAM                & \begin{tabular}[c]{@{}c@{}}NVIDIA \\ GPU(s)\end{tabular}                             & \begin{tabular}[c]{@{}c@{}}GPU Memory\\ (per card)\end{tabular}              \\ \hline
1       & 4/8                                                             & 16GB               & 1x 1050Ti                                                                            & 4GB GDDR5                                                                    \\ \hline
2       & 4/8                                                             & 32GB               & 1x GTX Titan                                                                         & 6GB GDDR5                                                                    \\ \hline
3       & 4/8                                                             & 16GB               & 1x RTX 2060                                                                          & 6GB GDDR6                                                                    \\ \hline
4       & 4/8                                                             & 32GB               & 2x RTX 2080Ti                                                                        & 11GB GDDR6                                                                   \\ \hline
5       & 4/8                                                             & 32GB               & \begin{tabular}[c]{@{}c@{}}1x RTX 2080Ti\\ 1x GTX Titan X\end{tabular}               & \begin{tabular}[c]{@{}c@{}}11GB GDDR6\\ 12GB GDDR5\end{tabular}              \\ \hline
6       & 4/8                                                             & 32GB               & 1x RTX 3090                                                                          & 24GB GDDR6X                                                                  \\ \hline
7       & 20/40                                                           & 40GB               & 2x V100                                                                              & 16GB HBM2                                                                    \\ \hline
8       & 20/40                                                           & 40GB               & 2x V100                                                                              & 16GB HBM2                                                                    \\ \hline
\multicolumn{5}{|c|}{Software}                                                                                                                                                                                                                                       \\ \hline
Machine & \begin{tabular}[c]{@{}c@{}}Operating\\ System\end{tabular}      & \multicolumn{3}{c|}{Other software}                                                                                                                                                      \\ \hline
1       & Ubuntu 20.04                                                    & \multicolumn{3}{c|}{\multirow{8}{*}{\begin{tabular}[c]{@{}c@{}}NVIDIA driver 470.57\\ NVIDIA CUDA Toolkit 11.4\\ \\ Python 3.8\\ \\ PyTorch 1.9 with \\ CUDA 11.1 support\end{tabular}}} \\ \cline{1-2}
2       & CentOS 7                                                        & \multicolumn{3}{c|}{}                                                                                                                                                                    \\ \cline{1-2}
3       & Ubuntu 20.04                                                    & \multicolumn{3}{c|}{}                                                                                                                                                                    \\ \cline{1-2}
4       & CentOS 7                                                        & \multicolumn{3}{c|}{}                                                                                                                                                                    \\ \cline{1-2}
5       & Ubuntu 20.04                                                    & \multicolumn{3}{c|}{}                                                                                                                                                                    \\ \cline{1-2}
6       & Ubuntu 20.04                                                    & \multicolumn{3}{c|}{}                                                                                                                                                                    \\ \cline{1-2}
7       & CentOS 7                                                        & \multicolumn{3}{c|}{}                                                                                                                                                                    \\ \cline{1-2}
8       & CentOS 7                                                        & \multicolumn{3}{c|}{}                                                                                                                                                                    \\ \hline
\end{tabular}
}
\caption{Machines used to train and inferencing multiple CNNs. All results in the following sections were obtained in these machines. All GPUs are from NVIDIA. Not all machines were available at the same time.}
\label{tab:machines_used}
\end{table}

To collect all the data we intended, we used multiple machines with a variety of hardware.
Table \ref{tab:machines_used} shows the hardware and software configurations we employed in our experiments.
These machines accumulated over 20000 hours of GPU computing, performing over 4500 small hyperparameter-tuning tests and over 2300 complete training tests. %
In the small tests, we trained \acp{cnn} for 30, 20 and 10 epochs on the MNIST, CIFAR-10 and ImageNet datasets, respectively.
The execution of these small tests was motivated by our observation that it is important to adopt optimal hyperparameters, initial learning rates and optimizer algorithms for each test, to ensure optimal model convergence. 
In the complete training tests, we trained the \acp{cnn} for 100, 200 and 100 epochs on the MNIST, CIFAR-10 and ImageNet datasets, respectively.
We trained all networks from scratch, i.e., we did not start the models with pre-trained weights.
Machines 7 and 8 were kindly provisioned by the \ac{lca} at University of Coimbra.
All hyperparameters used to obtain the final results and said results presented in this work can be found on our GitHub repository Wiki page \href{https://github.com/IT-Coimbra/RedBit/wiki/Quantization-Results}{"Quantization Results"}.
The optimizer used can be Adam or SGD. When SGD is used, momentum is set to 0.9. In both optimizers, weight decay is set to 0, except for the baseline tests, where weight decay was set to $0.0001$.

\subsection{Quantization Results}

We performed a series of tests using all quantization methods previously described to show how quantization affects \acp{cnn} applied to MNIST, CIFAR-10, and ImageNet.
Our goal is to evaluate different quantization methods and different levels of quantization, depending on the chosen bit width.

\subsubsection{MNIST}
Table \ref{tab:MNIST_LeNet5_ER} shows the best accuracy obtained in the validation set of MNIST with LeNet-5.
Each quantization method can offer different levels of quantization bit widths.
We were able to reproduce the results reported by the original works, as demonstrated by Table \ref{tab:MNIST-t-results}; the accuracy levels we observed are identical or superior.
We make three key observations.
First, QNN shows the biggest accuracy degradation when the LeNet-5 model is binarized.
The binarized model (W1A1) obtains 98.90\% accuracy compared to 99.69\% obtained in single-precision (W32A32).
If the input activations are quantized to 2 bits (W1A2), the quantized LeNet-5 model only gains 0.32\% of the 0.79\% accuracy loss between single-precision and the binarized model.
Second, DoReFa-Net offers a wide degree of quantization levels and suffers minimal accuracy loss
The quantized LeNet-5 model W1A2 using this quantization method achieves 99.61\% accuracy, representing an accuracy loss of just 0.06\% compared to single-precision.
Third, the LeNet-5 model can have its 32-bit weights binarized, corresponding to a reduction in of 32$\times$ in model size, while achieving minimal accuracy loss (less than 0.13\%).

\begin{table}[htb!]
\centering
\resizebox{\columnwidth}{!}{
\renewcommand{\tabcolsep}{2pt}
\begin{tabular}{|c|ccccccccc|}
\hline
\multirow{2}{*}{Method} & \multicolumn{9}{c|}{Quantization of Weights (W) and Input Activations (A)}                      \\
                                & W32A32  & W8A8    & W4A4    & W2A32     & W1A32   & W2A2    & W2A1    & W1A2    & W1A1    \\ \hline
QNN                             & 99.69\% & 99.52\% & 99.55\% & 99.63\%   & 99.56\% & 99.45\% & 98.88\% & 99.22\% & 98.90\% \\ \hline
DoReFa-Net                      & 99.67\% & 99.69\% & 99.68\% & 99.66\%   & 99.69\% & 99.62\% & 99.07\% & 99.61\% & 99.06\% \\ \hline
XNOR-Net                        & -       & -       & -       & -         & 99.63\% & -       & -       & -       & 99.37\% \\ \hline
TWN                             & 99.63\% & -       & -       & 99.63\% * & -       & -       & -       & -       & -       \\ \hline
TTQ                             & -       & -       & -       & 99.47\% * & -       & -       & -       & -       & -       \\ \hline
\end{tabular}
}
\caption{Results for the LeNet-5 network applied to MNIST.
Methods marked with a star (*) use ternary values, i.e., use 3 of 4 possible values with 2-bit representation; for quantization bit widths equal to 1, the quantization formula used is the \texttt{sign()} function instead of the quantization formula presented by the original work.
The baseline unmodified LeNet-5 CNN model achieves 99.59\% single-precision accuracy.}
\label{tab:MNIST_LeNet5_ER}
\end{table}

\begin{table}[t]
\centering
\resizebox{\columnwidth}{!}{
\renewcommand{\tabcolsep}{2pt}
\begin{tabular}{|c|ccccccccc|}
\hline
\multirow{2}{*}{Method} & \multicolumn{9}{c|}{Quantization of Weights (W) and Input Activations (A)}                      \\
                                & W32A32  & W8A8    & W4A4    & W2A32     & W1A32   & W2A2    & W2A1    & W1A2    & W1A1    \\ \hline
QNN                             & 91.06\% & 89.40\% & 89.29\% & 89.19\%   & 89.80\% & 83.14\% & 69.94\% & 82.73\% & 58.10\% \\ \hline
DoReFa-Net                      & 90.40\% & 89.42\% & 88.27\% & 89.56\%   & 89.42\% & 88.24\% & 62.17\% & 87.11\% & 62.70\% \\ \hline
XNOR-Net                        & -       & -       & -       & -         & 88.95\% & -       & -       & -       & 77.40\% \\ \hline
TWN                             & 90.59\% & -       & -       & 90.60\% * & -       & -       & -       & -       & -       \\ \hline
TTQ                             & -       & -       & -       & 89.11\% * & -       & -       & -       & -       & -       \\ \hline
\end{tabular}
}
\caption{Results for the ResNet-20 network applied to CIFAR-10. Methods marked with a star (*) use ternary values, i.e., use 3 of 4 possible values with 2-bit representation; for quantization bitwidths equal to 1, the quantization formula used is the \texttt{sign()} function instead of the normal quantization formula presented by the original work. The baseline unmodified ResNet-20 CNN model achieves 91.70\% single-precision accuracy.
\label{tab:CIFAR10_ResNet20_ER}}
\end{table}

\subsubsection{CIFAR-10}

Tables \ref{tab:CIFAR10_ResNet20_ER}, \ref{tab:CIFAR10_ResNet50_ER} and \ref{tab:CIFAR10_VGG16_ER} show the quatization results from the tests performed for the CIFAR-10 dataset using ResNet-20, ResNet-50 and VGG-16 models, respectively.

\noindent
\textbf{ResNet.} \Cref{tab:CIFAR10_ResNet20_ER,tab:CIFAR10_ResNet50_ER} show that increasing the depth from 20 to 50 layers improves accuracy results for almost all quantization scenarios.
These tables allow us to make five key observations.
First, QNN is the only quantization method that shows accuracy degradation when increasing the depth of ResNet.
It is also the method which offers the lowest accuracy when binarizing both weights and input activations on ResNet-20, with 58.10\% accuracy, compared to 77.40\% accuracy obtained by the binarized model using XNOR-Net.
Second, DoReFa-Net achieves 62.70\% and 63.46\% accuracy when binarizing both weights and input activations of ResNet-20 and ResNet-50 models, respectively. But it can achieve 87.11\% and 88.78\% when increasing the bit width of input activations to 2 bits, an accuracy loss of 3.29\% and 2.86\% when compared to single-precision accuracy of ResNet-20 and -50 models, respectively.
Third, columns W2A2, W2A1 and W1A2 show that input activations are more important than weights. The impact of lower bit widths of the input activations on accuracy is greater than that of the weights.
Fourth, binarizing weights but keeping input activations in single-precision reduces the size of the quantized models by up to 32$\times$ while incurring less than 2.11\% and 2.23\% accuracy loss for ResNet-20 and -50 models, respectively.
Finally, increasing the bit width to 2 bits or using ternary values for the weights while keeping the input activations in single-precision has little to no effect on accuracy gain when compared to the previously mentioned results, W1A32.

\begin{table}[h]
\centering
\resizebox{\columnwidth}{!}{
\renewcommand{\tabcolsep}{2pt}
\begin{tabular}{|c|ccccccccc|}
\hline
\multirow{2}{*}{Method} & \multicolumn{9}{c|}{Quantization of Weights (W) and Input Activations (A)}                      \\
                                & W32A32  & W8A8    & W4A4    & W2A32     & W1A32   & W2A2    & W2A1    & W1A2    & W1A1    \\ \hline
QNN                             & 92.67\% & 88.83\% & 88.66\% & 91.42\%   & 91.66\% & 82.92\% & 72.99\% & 81.93\% & DNC     \\ \hline
DoReFa-Net                      & 91.64\% & 89.46\% & 89.73\% & 91.26\%   & 91.37\% & 89.11\% & 68.61\% & 88.78\% & 63.46\% \\ \hline
XNOR-Net                        & -       & -       & -       & -         & 90.44\% & -       & -       & -       & 81.90\% \\ \hline
TWN                             & 92.45\% & -       & -       & 91.89\% * & -       & -       & -       & -       & -       \\ \hline
TTQ                             & -       & -       & -       & 90.91\% * & -       & -       & -       & -       & -       \\ \hline
\end{tabular}
}
\caption{Results for the ResNet-50 network applied to CIFAR-10. Methods marked with a star (*) use ternary values, i.e., use 3 of 4 possible values with 2-bit representation; DNC - Does Not Converge; For quantization bitwidths equal to 1, the quantization formula used is the \texttt{sign()} function instead of the normal quantization formula presented by the original work. The baseline unmodified ResNet-50 CNN model achieves 92.97\% single-precision accuracy.}
\label{tab:CIFAR10_ResNet50_ER}
\end{table}

\noindent
\textbf{VGG.}
Table \ref{tab:CIFAR10_VGG16_ER} shows the results obtained with VGG-16.
We make five key observations. 
First, VGG-16 achieves better results for single-precision than the ResNet models.
Second, VGG-16 achieves the best result when binarizing the weights but keeping the input activations in single-precision with the DoReFa-Net quantization method.
In this case, the model achieves 92.67\%, a degradation of 0.16\% compared to the accuracy using single-precision weights. 
Third, depending on the adopted bit width quantization, the model may struggle to converge on an optimal solution. 
Fourth, DoReFa-Net achieves a good result for binary weights and 2-bit quantization of input activations.
It achieves an accuracy of 90.12\%, corresponding to a degradation of 2.71\%.
Finally, our results for the CIFAR-10 dataset are in line with what is expected when compared to the state-of-the-art single-precision accuracy results shown in Table \ref{tab:CIFAR10-t-results}.

\begin{table}[t]
\centering
\resizebox{\columnwidth}{!}{
\renewcommand{\tabcolsep}{2pt}
\begin{tabular}{|c|ccccccccc|}
\hline
\multirow{2}{*}{Method} & \multicolumn{9}{c|}{Quantization of Weights (W) and Activations (A)}                      \\
                                & W32A32  & W8A8    & W4A4    & W2A32     & W1A32   & W2A2    & W2A1    & W1A2    & W1A1    \\ \hline
QNN                             & 92.97\% & 88.40\% & 88.13\% & 91.99\%   & DNC     & 86.45\% & 35.99\% & DNC     & DNC     \\ \hline
DoReFa-Net                      & 92.83\% & 70.39\% & 69.43\% & 90.84\%   & 92.67\% & 68.24\% & DNC     & 90.12\% & 73.96\% \\ \hline
XNOR-Net                        & -       & -       & -       & -         & 92.23\% & -       & -       & -       & 77.98\% \\ \hline
TWN                             & 92.93\% & -       & -       & 92.30\% * & -       & -       & -       & -       & -       \\ \hline
TTQ                             & -       & -       & -       & DNC       & -       & -       & -       & -       & -       \\ \hline
\end{tabular}
}
\caption{Results for the VGG-16 network applied to CIFAR-10. Methods marked with a star (*) use ternary values, i.e., use 3 of 4 possible values with 2-bit representation; DNC - Does Not Converge; For quantization bit-widths equal to 1, the quantization formula used is the \texttt{sign()} function instead of the normal quantization formula presented by the original work. The baseline unmodified VGG-16 CNN model achieves 93.37\% single-precision accuracy.}
\label{tab:CIFAR10_VGG16_ER}
\end{table}

\begin{table}[b]
\centering
\resizebox{\columnwidth}{!}{
\renewcommand{\tabcolsep}{2pt}
\begin{tabular}{|c|c|ccccccccc|}
\hline
\multicolumn{2}{|c|}{\multirow{2}{*}{Method}}                      & \multicolumn{9}{c|}{Quantization of Weights (W) and Activations (A)}                                                                                                                                                                                                                                                                                                                                                                                                                                                                                        \\ \cline{3-11} 
\multicolumn{2}{|c|}{}                                             & W32A32                                                    & W8A8                                                      & W4A4                                                      & W2A32                                                       & W1A32                                                     & W2A2                                                      & W2A1                                                      & W1A2                                                      & W1A1                                                      \\ \hline
QNN        & \begin{tabular}[c]{@{}c@{}}Top-1\\ Top-5\end{tabular} & \begin{tabular}[c]{@{}c@{}}59.26\%\\ 81.56\%\end{tabular} & \begin{tabular}[c]{@{}c@{}}55.48\%\\ 78.39\%\end{tabular} & \begin{tabular}[c]{@{}c@{}}53.18\%\\ 76.62\%\end{tabular} & \begin{tabular}[c]{@{}c@{}}46.44\%\\ 71.19\%\end{tabular}   & \begin{tabular}[c]{@{}c@{}}52.57\%\\ 76.26\%\end{tabular} & \begin{tabular}[c]{@{}c@{}}37.09\%\\ 61.91\%\end{tabular} & \begin{tabular}[c]{@{}c@{}}29.51\%\\ 53.53\%\end{tabular} & \begin{tabular}[c]{@{}c@{}}45.63\%\\ 70.01\%\end{tabular} & \begin{tabular}[c]{@{}c@{}}38.42\%\\ 62.96\%\end{tabular} \\ \hline
DoReFa-Net & \begin{tabular}[c]{@{}c@{}}Top-1\\ Top-5\end{tabular} & \begin{tabular}[c]{@{}c@{}}53.39\%\\ 72.90\%\end{tabular} & \begin{tabular}[c]{@{}c@{}}50.52\%\\ 70.56\%\end{tabular} & \begin{tabular}[c]{@{}c@{}}50.35\%\\ 70.92\%\end{tabular} & \begin{tabular}[c]{@{}c@{}}51.10\%\\ 71.67\%\end{tabular}   & \begin{tabular}[c]{@{}c@{}}50.93\%\\ 71.47\%\end{tabular} & \begin{tabular}[c]{@{}c@{}}48.95\%\\ 70.38\%\end{tabular} & \begin{tabular}[c]{@{}c@{}}30.82\%\\ 53.80\%\end{tabular} & \begin{tabular}[c]{@{}c@{}}47.96\%\\ 70.17\%\end{tabular} & \begin{tabular}[c]{@{}c@{}}30.40\%\\ 53.29\%\end{tabular} \\ \hline
XNOR-Net   & \begin{tabular}[c]{@{}c@{}}Top-1\\ Top-5\end{tabular} & -                                                         & -                                                         & -                                                         & -                                                           & \begin{tabular}[c]{@{}c@{}}52.06\%\\ 74.89\%\end{tabular} & -                                                         & -                                                         & -                                                         & \begin{tabular}[c]{@{}c@{}}42.53\%\\ 66.41\%\end{tabular} \\ \hline
TWN        & \begin{tabular}[c]{@{}c@{}}Top-1\\ Top-5\end{tabular} & \begin{tabular}[c]{@{}c@{}}59.22\%\\ 81.60\%\end{tabular} & -                                                         & -                                                         & \begin{tabular}[c]{@{}c@{}}55.36\%*\\ 78.61\%*\end{tabular} & -                                                         & -                                                         & -                                                         & -                                                         & -                                                         \\ \hline
TTQ        & \begin{tabular}[c]{@{}c@{}}Top-1\\ Top-5\end{tabular} & -                                                         & -                                                         & -                                                         & \begin{tabular}[c]{@{}c@{}}43.17\%*\\ 68.07\%*\end{tabular} & -                                                         & -                                                         & -                                                         & -                                                         & -                                                         \\ \hline
\end{tabular}
}
\caption{Accuracy results (Top-1; Top-5) for AlexNet trained with ImageNet. Methods marked with a star (*) use ternary values, i.e., use 3 of 4 possible values with 2 bit representation; for quantization bit widths equal to 1, the quantization formula used is the \texttt{sign()} function instead of the normal quantization formula presented by the original work. The baseline unmodified AlexNet CNN model achieves 61.01\% top-1 and 82.95\% top-5 single-precision accuracy.}
\label{tab:ImageNet_AlexNet_ER}
\end{table}

\gboxbegin{kobackbright}{koborderbright}{OBSERVATION}{\rtaskobs{ko}}
The following general key observation can be derived: across all evaluated models, the hyperparameter search for the best optimizer and initial learning rate is crucial to maximize the accuracy.
\gboxend

\subsubsection{ImageNet}
For the ImageNet dataset, Tables \ref{tab:ImageNet_AlexNet_ER}, \ref{tab:ImageNet_ResNet18_ER} and \ref{tab:ImageNet_VGG16_ER} gather the best accuracy results achieved with AlexNet, ResNet-18, and VGG-16, respectively.

\gboxbegin{kobackbright}{koborderbright}{OBSERVATION}{\rtaskobs{ko}}
The quantization method of DoReFa-Net~\cite{dorefa-net} offers both a wide variety of quantization levels and more consistent results;
\gboxend

\noindent
\textbf{AlexNet.}
Table \ref{tab:ImageNet_AlexNet_ER} shows the results obtained with AlexNet.
We make four key observations.
First, the discrepancy in accuracy results between quantization methods in single-precision is justified by the architecture differences applied by each method on the baseline AlexNet \ac{cnn} model.
Second, \ac{twn} achieves 55.36\% top-1 and 78.61\% top-5 when using ternary values for the weights and single-precision input activations. An accuracy degradation of 3.86\% top-1 and 2.99\% top-5 compared to single-precision weights.
Third, keeping the input activations in single-precision but binarizing the weights (W1A32), \ac{qnn}, DoReFa-Net and \ac{bwn} achieve 52.57\% top-1 and 76.26\% top-5, 50.93\% top-1 and 71.47\% top-5 and 52.06\% top-1 and 74.89\% top-5, respectively. These models maintain an acceptable accuracy while using binary weights.
Fourth, \ac{qnn} achieves 38.42\% top-1 and 62.96\% accuracy when binarizing both weights and input activations incurring in 20.84\% and 18.60\% accuracy loss for top-1 and top-5, respectively, compared to single-precision. XNOR-Net, by using scaling factors, not binarizing the first and last convolutional/fully connected layers and using layer reordering, increases the accuracy slightly to 42.53\% top-1 and 66.41\%.

\noindent
\textbf{ResNet.} 
Table \ref{tab:ImageNet_ResNet18_ER} shows the results obtained with ResNet-18.
We make three key observations.
First, DoReFa-Net is a stable quantization method, offering good performance accross all quantization levels. With DoReFa-Net, is possible to binarize the weights and keep input activations in single-precision, achieving 62.13\% top-1 and 83.66\% top-5 accuracy, incurring 0.29\% and 0.51\% accuracy degradation for top-1 and top-5, respectively, compared to single-precision.
Second, looking at columns W2A2, W2A1 and W1A2, is evident that input activations have a greater effect on accuracy than weights. If input activations are kept with higher bit width, is possible to achieve better accuracies.
Third, XNOR-Net continues to be the best quantization method to binarize both weights and input activations, achieving 49.26\% top-1 and 73.53\% top-5 accuracy.

\gboxbegin{kobackbright}{koborderbright}{OBSERVATION}{\rtaskobs{ko}}
The training of quantized \ac{bwn} and XNOR-Net~\cite{xnor-net} models revealed that the application of the ReLU non-linearity after the convolution operations yields better results. We apply this optimization in all \ac{bwn} and XNOR-Net results we report;
\gboxend

\begin{table}[htb!]
\centering
\resizebox{\columnwidth}{!}{
\renewcommand{\tabcolsep}{2pt}
\begin{tabular}{|c|c|ccccccccc|}
\hline
\multicolumn{2}{|c|}{\multirow{2}{*}{Paper / Method}}              & \multicolumn{9}{c|}{Quantization of Weights (W) and Activations (A)}                                                                                                                                                                                                                                                                                                                                                                                                                                                                                         \\ \cline{3-11} 
\multicolumn{2}{|c|}{}                                             & W32A32                                                    & W8A8                                                      & W4A4                                                      & W2A32                                                       & W1A32                                                     & W2A2                                                       & W2A1                                                      & W1A2                                                      & W1A1                                                      \\ \hline
QNN        & \begin{tabular}[c]{@{}c@{}}Top-1\\ Top-5\end{tabular} & \begin{tabular}[c]{@{}c@{}}62.54\%\\ 84.45\%\end{tabular} & \begin{tabular}[c]{@{}c@{}}61.43\%\\ 83.31\%\end{tabular} & \begin{tabular}[c]{@{}c@{}}60.31\%\\ 82.67\%\end{tabular} & \begin{tabular}[c]{@{}c@{}}53.46\%\\ 77.76\%\end{tabular}   & \begin{tabular}[c]{@{}c@{}}59.06\%\\ 82.01\%\end{tabular} & \begin{tabular}[c]{@{}c@{}}38.00\%;\\ 63.82\%\end{tabular} & DNC                                                       & \begin{tabular}[c]{@{}c@{}}32.95\%\\ 58.57\%\end{tabular} & \begin{tabular}[c]{@{}c@{}}15.18\%\\ 34.75\%\end{tabular} \\ \hline
DoReFa-Net & \begin{tabular}[c]{@{}c@{}}Top-1\\ Top-5\end{tabular} & \begin{tabular}[c]{@{}c@{}}62.42\%\\ 84.17\%\end{tabular} & \begin{tabular}[c]{@{}c@{}}61.38\%\\ 83.29\%\end{tabular} & \begin{tabular}[c]{@{}c@{}}61.11\%\\ 83.21\%\end{tabular} & \begin{tabular}[c]{@{}c@{}}62.09\%\\ 83.90\%\end{tabular}   & \begin{tabular}[c]{@{}c@{}}62.13\%\\ 83.66\%\end{tabular} & \begin{tabular}[c]{@{}c@{}}60.73\%\\ 82.99\%\end{tabular}  & \begin{tabular}[c]{@{}c@{}}26.58\%\\ 49.24\%\end{tabular} & \begin{tabular}[c]{@{}c@{}}59.93\%\\ 82.03\%\end{tabular} & \begin{tabular}[c]{@{}c@{}}41.17\%\\ 65.77\%\end{tabular} \\ \hline
XNOR-Net   & \begin{tabular}[c]{@{}c@{}}Top-1\\ Top-5\end{tabular} & -                                                         & -                                                         & -                                                         & -                                                           & \begin{tabular}[c]{@{}c@{}}61.24\%\\ 83.33\%\end{tabular} & -                                                          & -                                                         & -                                                         & \begin{tabular}[c]{@{}c@{}}49.26\%\\ 73.53\%\end{tabular} \\ \hline
TWN        & \begin{tabular}[c]{@{}c@{}}Top-1\\ Top-5\end{tabular} & \begin{tabular}[c]{@{}c@{}}62.70\%\\ 84.46\%\end{tabular} & -                                                         & -                                                         & \begin{tabular}[c]{@{}c@{}}60.59\%*\\ 82.80\%*\end{tabular} & -                                                         & -                                                          & -                                                         & -                                                         & -                                                         \\ \hline
TTQ        & \begin{tabular}[c]{@{}c@{}}Top-1\\ Top-5\end{tabular} & -                                                         & -                                                         & -                                                         & \begin{tabular}[c]{@{}c@{}}51.62\%*\\ 74.85\%*\end{tabular} & -                                                         & -                                                          & -                                                         & -                                                         & -                                                         \\ \hline
\end{tabular}
}
\caption{Accuracy results (Top-1; Top-5) for ResNet-18 trained with ImageNet. Methods marked with a star (*) use ternary values, i.e., use 3 of 4 possible values with 2 bit representation; DNC - Does Not Converge; For quantization bit widths equal to 1, the quantization formula used is the \texttt{sign()} function instead of the normal quantization formula presented by the original work. The baseline unmodified ResNet-18 CNN model achieves 67.05\% top-1 and 87.66\% top-5 single-precision accuracy.}
\label{tab:ImageNet_ResNet18_ER}
\end{table}

\noindent
\textbf{VGG.}
Table \ref{tab:ImageNet_VGG16_ER} shows results of VGG-16 trained with ImageNet. \textbf{These results were obtained with only 10 epochs of training}.
We make five key observations.
First, even with only 10 epochs trained, VGG-16 is capable of achieving better results than AlexNet in single precision. Second, for DoReFa-Net, is possible to reduce the VGG-16 model size by up to 32$\times$ by binarizing the weights and keeping the input activations in single precision and still achieve 57.65\% top-1 and 81.53\% top-5 accuracy, a 3.15\% and 2.19\% accuracy loss for top-1 and top-5, respectively, compared to single-precision. DoReFa-Net is unstable when quantizing the weights to bit widths different than 1. XNOR-Net also achieves similar results with 56.22\% top-1 and 80.38\% top-5 accuracy, when only the weights are quantized to 1 bit (\ac{bwn}). Third, \ac{twn} achieves 57.87\% top-1 and 81.78\% while using ternary weight values, an accuracy loss compared to single-precision of 1.38\% top-1 and 1.30\% top-5. Fourth, when both weights and input activations are binarized, the accuracy loss is significant. DoReFa-Net is the quantization method to get better results here, 29.92\% top-1 and 54.74\% top-5 accuracy, still a significant 30.88\% and 28.98\% accuracy loss for top-1 and top-5, respectively, compared to single-precision. Fifth, increasing the input activations bit width to 2 bits while keeping the weights binarized (W1A2), increases the accuracy. DoReFa-Net now achieves 53.83\% top-1 and 78.35\% top-5 accuracy.

\begin{table}[t]
\centering
\resizebox{\columnwidth}{!}{
\renewcommand{\tabcolsep}{2pt}
\begin{tabular}{|c|c|ccccccccc|}
\hline
\multicolumn{2}{|c|}{\multirow{2}{*}{Method}}                      & \multicolumn{9}{c|}{Quantization of Weights (W) and Activations (A)}                                                                                                                                                                                                                                                                                                                                                                                                                                                                                          \\ \cline{3-11} 
\multicolumn{2}{|c|}{}                                             & W32A32                                                    & W8A8                                                      & W4A4                                                      & W2A32                                                         & W1A32                                                     & W2A2                                                      & W2A1                                                      & W1A2                                                      & W1A1                                                      \\ \hline
QNN        & \begin{tabular}[c]{@{}c@{}}Top-1\\ Top-5\end{tabular} & \begin{tabular}[c]{@{}c@{}}60.98\%\\ 83.77\%\end{tabular} & \begin{tabular}[c]{@{}c@{}}45.15\%\\ 70.37\%\end{tabular} & \begin{tabular}[c]{@{}c@{}}44.19\%\\ 69.66\%\end{tabular} & \begin{tabular}[c]{@{}c@{}}47.01\%\\ 72.76\%\end{tabular}     & DNC                                                       & \begin{tabular}[c]{@{}c@{}}40.59\%\\ 66.00\%\end{tabular} & \begin{tabular}[c]{@{}c@{}}31.21\%\\ 55.85\%\end{tabular} & DNC                                                       & DNC                                                       \\ \hline
DoReFa-Net & \begin{tabular}[c]{@{}c@{}}Top-1\\ Top-5\end{tabular} & \begin{tabular}[c]{@{}c@{}}60.80\%\\ 83.72\%\end{tabular} & DNC                                                       & DNC                                                       & DNC                                                           & \begin{tabular}[c]{@{}c@{}}57.65\%\\ 81.53\%\end{tabular} & DNC                                                       & DNC                                                       & \begin{tabular}[c]{@{}c@{}}53.83\%\\ 78.35\%\end{tabular} & \begin{tabular}[c]{@{}c@{}}29.92\%\\ 54.74\%\end{tabular} \\ \hline
XNOR-Net   & \begin{tabular}[c]{@{}c@{}}Top-1\\ Top-5\end{tabular} & -                                                         & -                                                         & -                                                         & -                                                             & \begin{tabular}[c]{@{}c@{}}56.22\%\\ 80.38\%\end{tabular} & -                                                         & -                                                         & -                                                         & \begin{tabular}[c]{@{}c@{}}16.48\%\\ 38.08\%\end{tabular} \\ \hline
TWN        & \begin{tabular}[c]{@{}c@{}}Top-1\\ Top-5\end{tabular} & \begin{tabular}[c]{@{}c@{}}59.25\%\\ 83.08\%\end{tabular} & -                                                         & -                                                         & \begin{tabular}[c]{@{}c@{}}57.87\% *\\ 81.78\% *\end{tabular} & -                                                         & -                                                         & -                                                         & -                                                         & -                                                         \\ \hline
TTQ        & \begin{tabular}[c]{@{}c@{}}Top-1\\ Top-5\end{tabular} & -                                                         & -                                                         & -                                                         & DNC                                                             & -                                                         & -                                                         & -                                                         & -                                                         & -                                                         \\ \hline
\end{tabular}
}
\caption{Accuracy results (Top-1; Top-5) for VGG-16 network trained with ImageNet (Results after 10 epochs). Methods marked with a star (*) use ternary values, i.e., use 3 of 4 possible values with 2-bit representation; for quantization bit widths equal to 1, the quantization formula used is the \texttt{sign()} function instead of the normal quantization formula presented by the original work. DNC - Does Not Converge. The baseline unmodified VGG-16 network achieves 60.82\% top-1 and 84.00\% top-5 accuracy.}
\label{tab:ImageNet_VGG16_ER}
\end{table}

A holistic analysis of the above results allows us to derive the network-specific key observations regarding quantization, identified in boxes KEY OBESRVATION 1 to 4.

\gboxbegin{kobackbright}{koborderbright}{OBSERVATION}{\rtaskobs{ko}}
TWN~\cite{twn} is a good quantization method. It offers consistent results and is overall the best method to quantize to the W2A32 quantization level.
\gboxend

\begin{figure*}[t]
\centering
\begin{tabular}{ccc}

\subcaptionbox{\label{subfig:mnist-lenet5-quant-weights-activations}}{\includegraphics[width = 56mm]{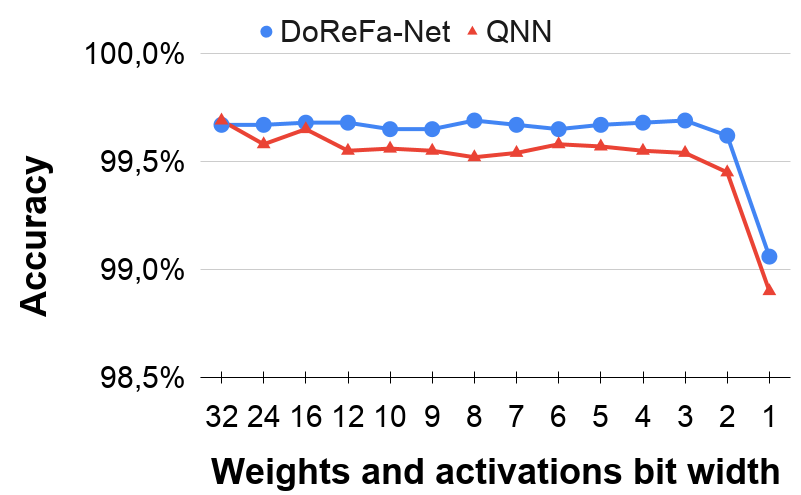}} &

\subcaptionbox{\label{subfig:mnist-lenet5-quant-weights-A32}}{\includegraphics[width = 56mm]{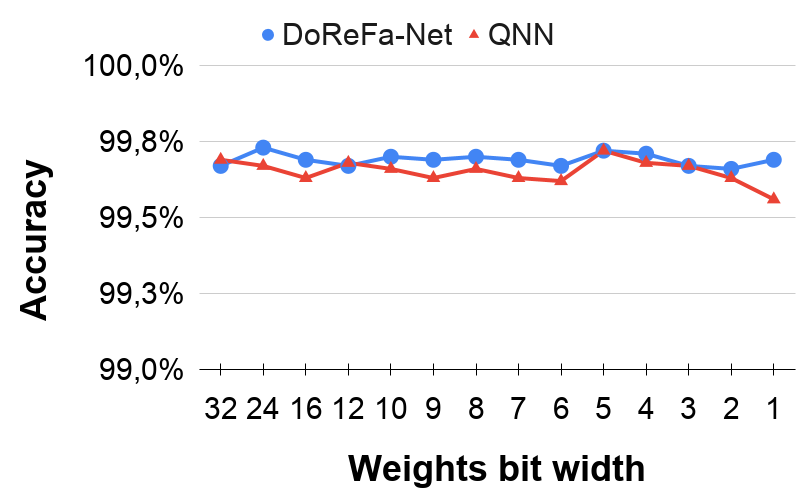}} &

\subcaptionbox{\label{subfig:mnist-lenet5-quant-activations-W1}}{\includegraphics[width = 56mm]{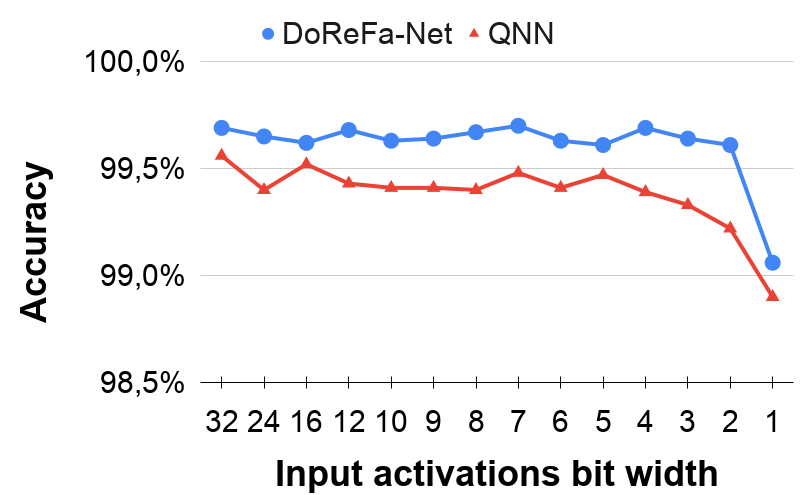}}\\

\subcaptionbox{\label{subfig:cifar10-resnet20-quant-weights-activations}}{\includegraphics[width = 56mm]{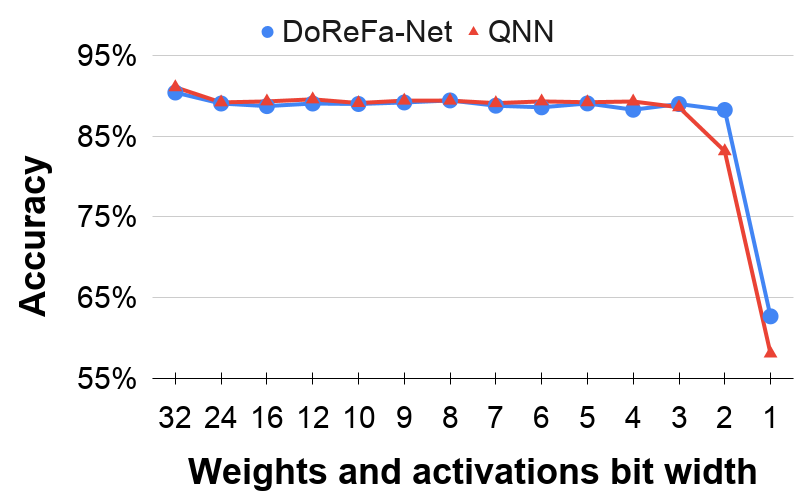}} &

\subcaptionbox{\label{subfig:cifar10-resnet20-quant-weights-A32}}{\includegraphics[width = 56mm]{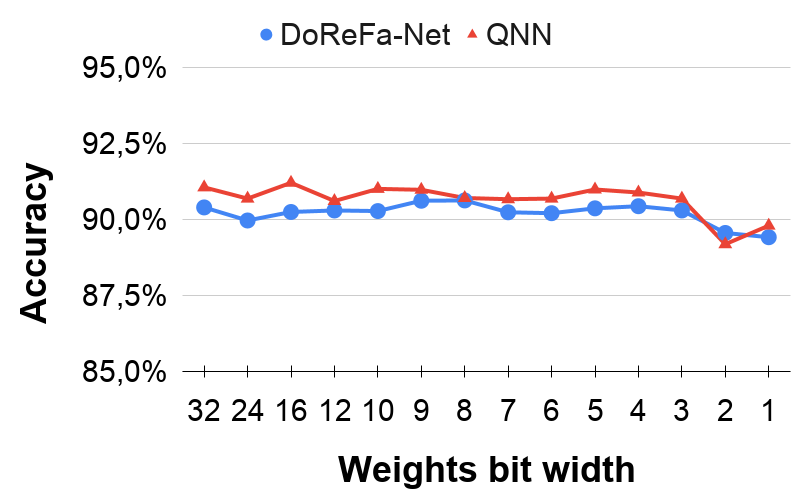}} &

\subcaptionbox{\label{subfig:cifar10-resnet20-quant-activations-W1}}{\includegraphics[width = 56mm]{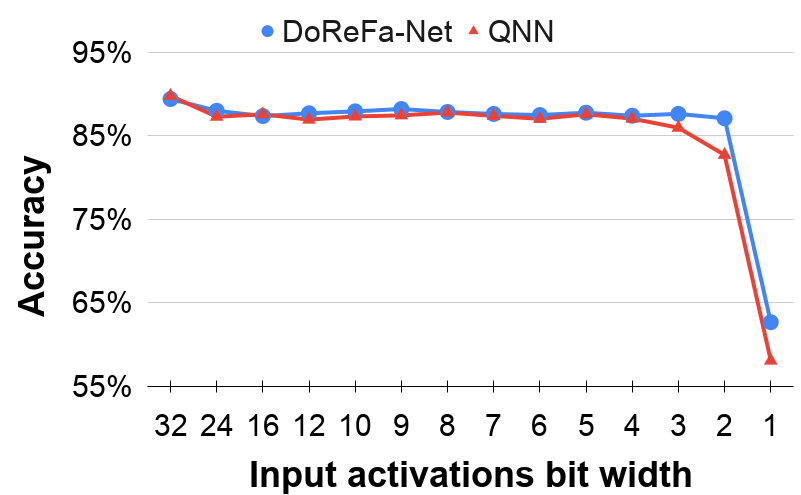}}

\end{tabular}
\caption{Quantization curves for LeNet-5 trained with MNIST and ResNet-20 trained with CIFAR-10.
\textbf{(a)} Quantization of weights in LeNet-5. Input activations are in single precision.
\textbf{(b)} Quantization of weights and input activations in LeNet-5.
\textbf{(c)} Quantization of input activations in LeNet-5. Weights are binarized.
\textbf{(d)} Quantization of weights in ResNet-20. Input activations are in single precision.
\textbf{(e)} Quantization of weights and input activations in ResNet-20.
\textbf{(f)} Quantization of input activations in ResNet-20. Weights are binarized.
}
\label{fig:plots_2d}
\end{figure*}

\section{Design Space Exploration}
\label{dse}

\begin{figure*}[t]
\centering
\begin{tabular}{cccc}
\subcaptionbox{\label{subfig:lenet5-qnn}}{\includegraphics[width=0.23\textwidth]{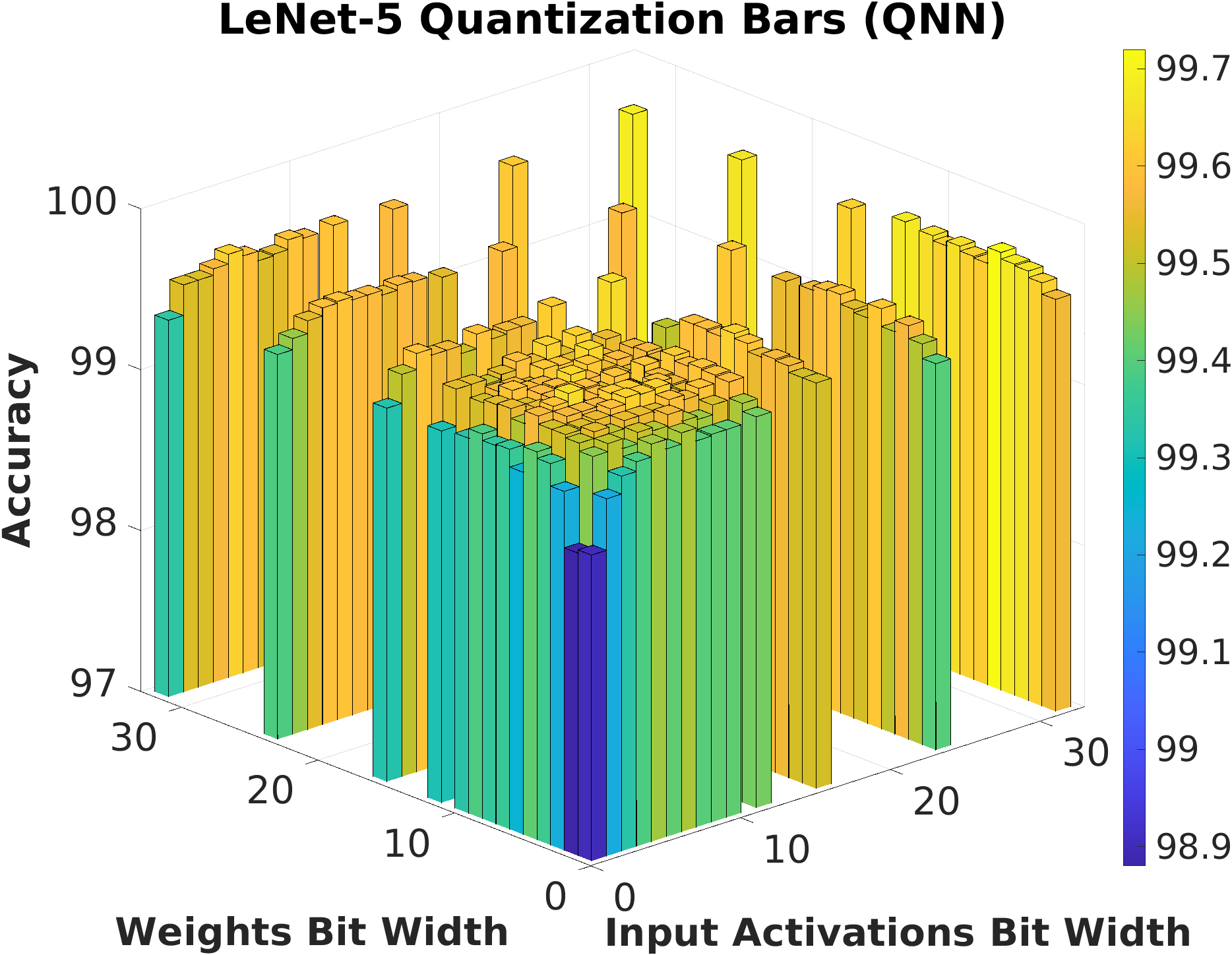}} &
\subcaptionbox{\label{subfig:lenet5-dorefa-net}}{\includegraphics[width=0.23\textwidth]{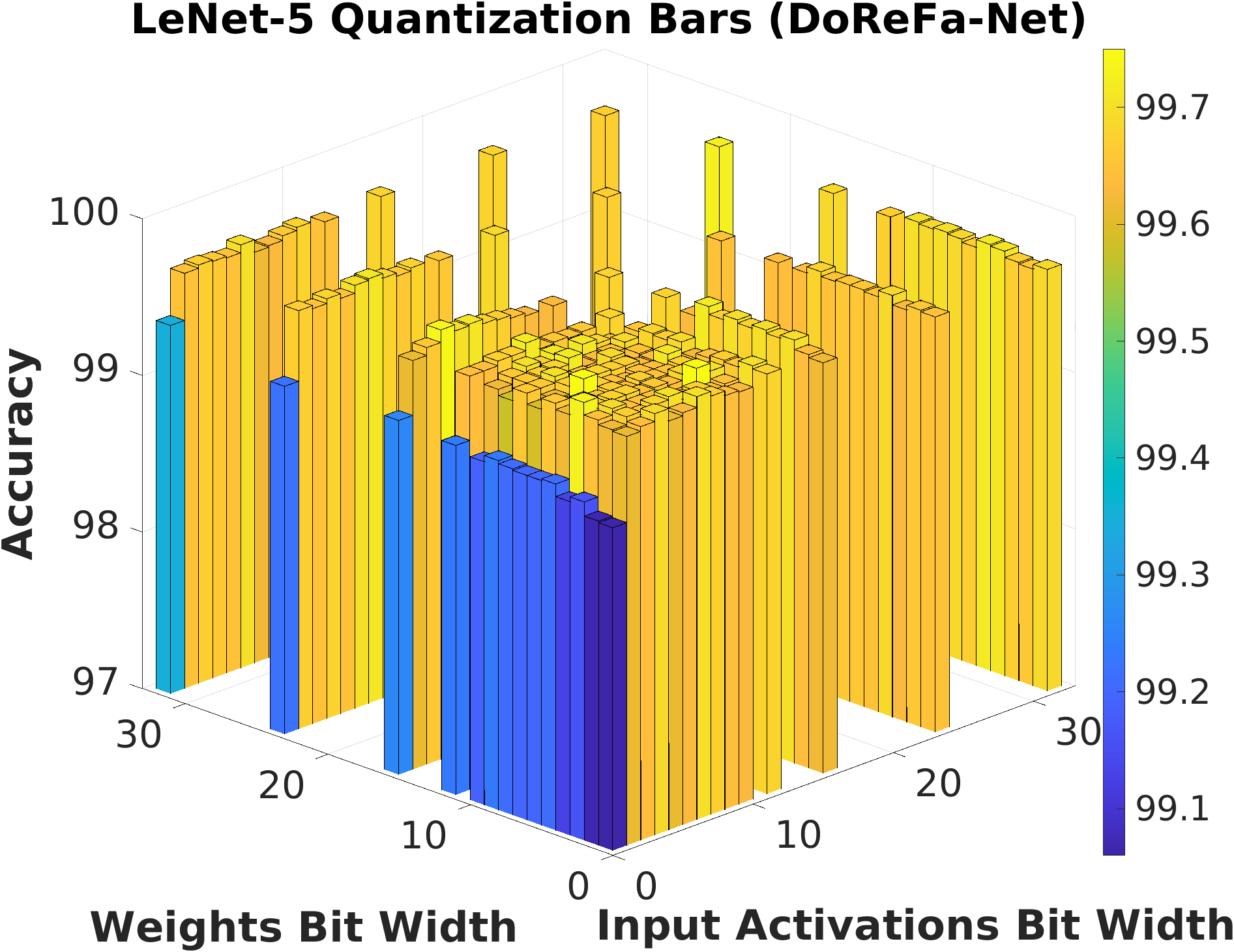}} &
\subcaptionbox{\label{subfig:resnet20-qnn}}{\includegraphics[width=0.23\textwidth]{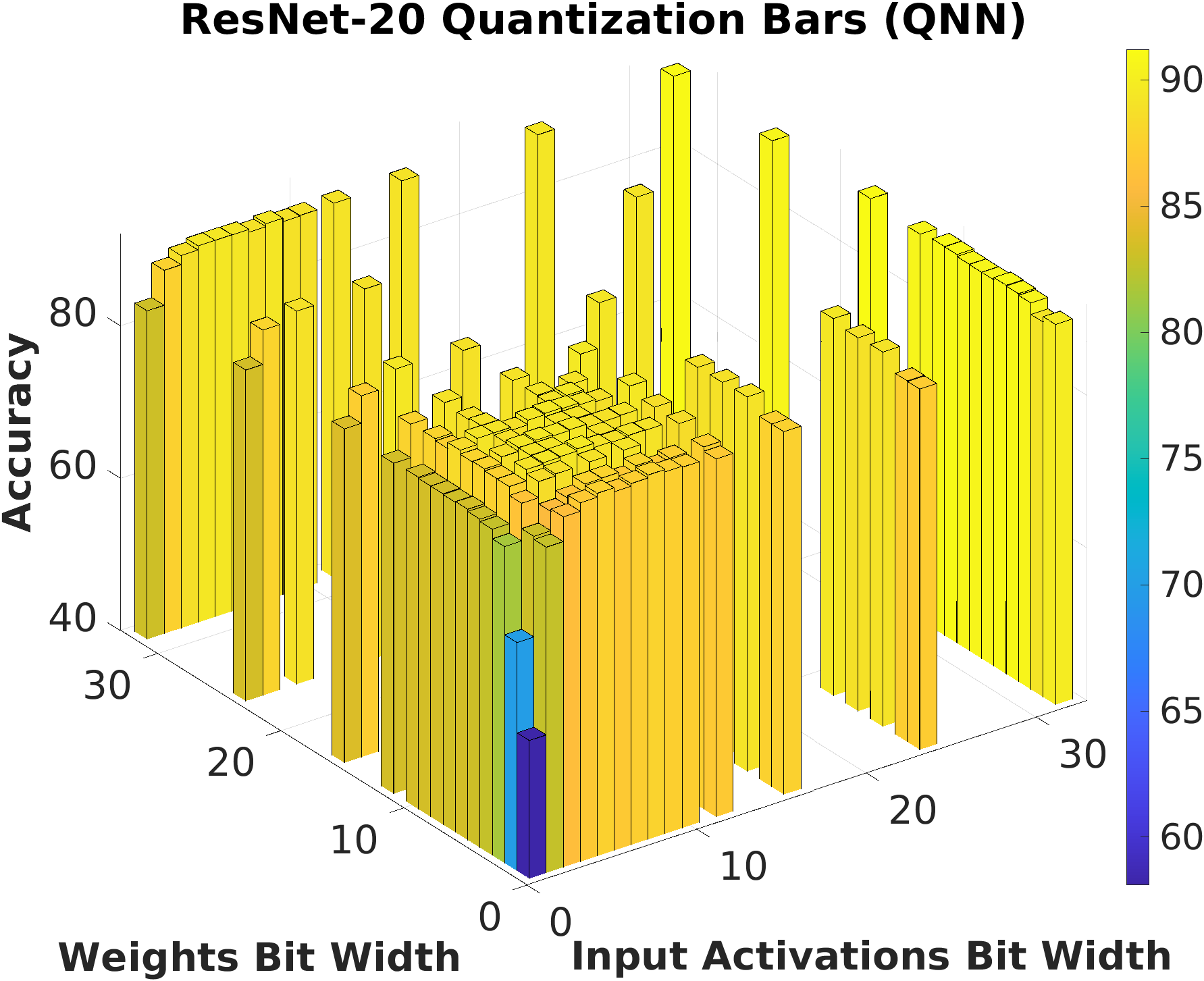}} &
\subcaptionbox{\label{subfig:resnet20-dorefa-net}}{\includegraphics[width=0.23\textwidth]{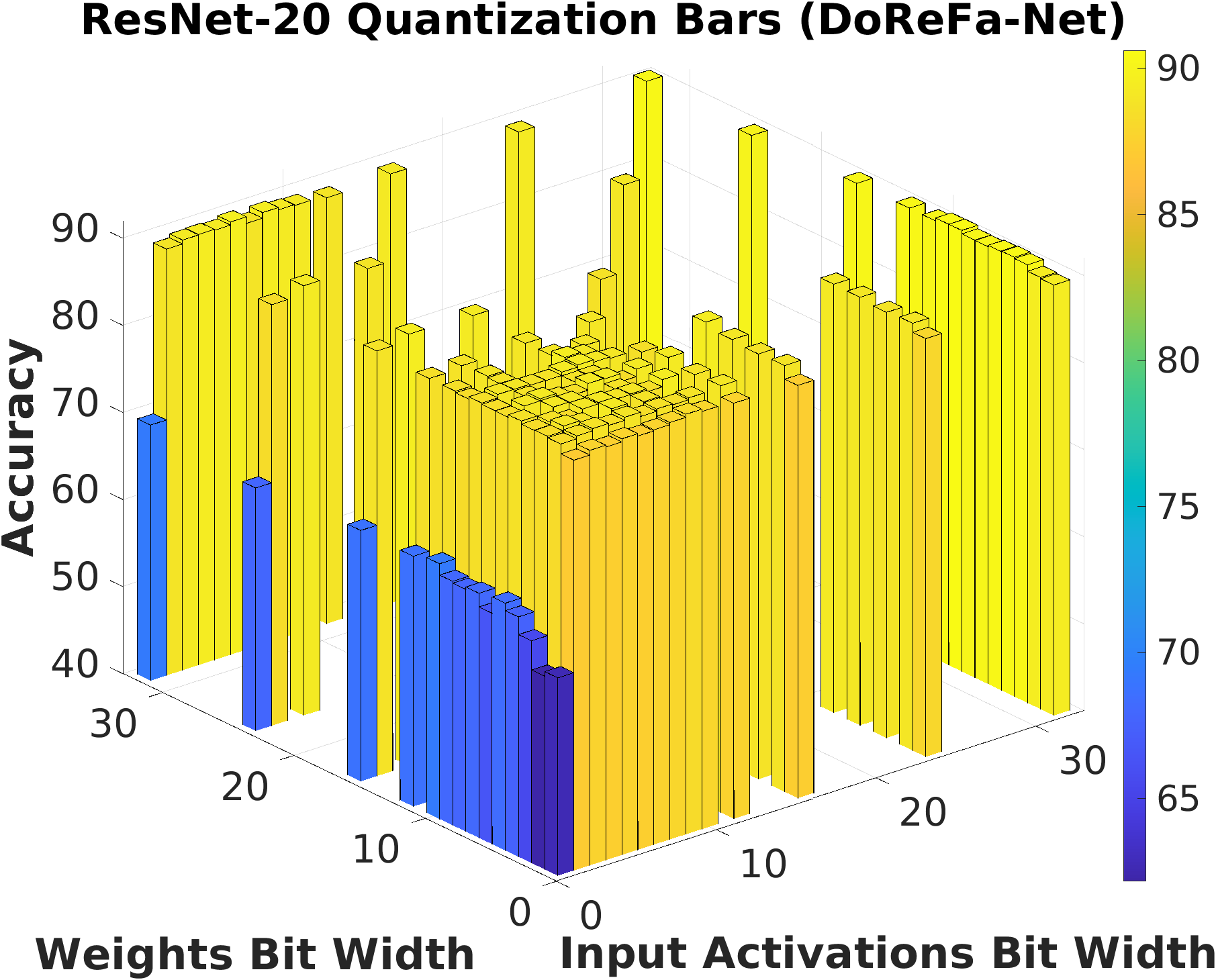}}
\end{tabular}
\caption{3D scatterbar of quantization applied to LeNet-5 and ResNet-20, trained with MNIST and CIFAR-10, respectively, with QNN and DoReFa-Net methods. \textbf{(a)} QNN Quantization of LeNet-5. \textbf{(b)} DoReFa-Net Quantization of LeNet-5. \textbf{(c)} QNN Quantization of ResNet-20. \textbf{(d)} DoReFa-Net Quantization of ResNet-20.
}
\label{fig:plots_3d}
\end{figure*}

Since both DoReFa-Net and QNN offer a wide degree of quantization levels, we trained a wide variety of quantized models and represent these data in \Cref{fig:plots_2d,fig:plots_3d}.
The quantization methods used to obtain these results were DoReFa-Net and QNN applied to LeNet-5 (on the MNIST dataset) and to ResNet-20 (on the CIFAR-10 dataset).
The 3D representation is made with a bar chart, where each bar represents the accuracy obtained with the quantized model to the desired bit width for weights and input activations.

\subsection{Reducing Computational Cost and Model Size Without Compromising Accuracy}
\label{sec:reducing-comp-cost}

To reduce the computational cost of arithmetic operations, both the bit widths of weights and input activations can be reduced.
\Cref{subfig:mnist-lenet5-quant-weights-activations,subfig:cifar10-resnet20-quant-weights-activations} show the accuracy curves when quantizing both weights and input activations to the desired bit width.
Overall, both methods behave similarly, giving good results until 2-bit quantization.
Binarization continues to be a challenge.
In the case of LeNet-5 trained with MNIST using the DoReFa-Net method, the \ac{cnn} model can be quantized from 32 bits down to 2 bits with a 0.07\% accuracy loss.
In the case of QNN, the model can only be quantized from 32 bits down to 3 bits with an accuracy loss of 0.17\%.
For ResNet-20 trained with CIFAR-10, quantizing from 32 bits down to 2 bits, DoReFa-Net achieves an accuracy loss less than 2.2\%, with an average of 88.97\% accuracy.
QNN can only present an accuracy degradation less than 2.5\% when quantizing from 32 bits down to 3 bits.
Lower than 3 bits, QNN incurs higher accuracy degradation than DoReFa-Net.
\Cref{subfig:mnist-lenet5-quant-weights-A32,subfig:cifar10-resnet20-quant-weights-A32} show that it is possible to achieve minimal accuracy losses (of less than $1$\%) with binarized weights. This corresponds to a model size reduction of up to 32$\times$, which provides savings in computational costs by reducing the bit width of the input activations.
\Cref{subfig:mnist-lenet5-quant-activations-W1,subfig:cifar10-resnet20-quant-activations-W1} show the accuracy curves when the weights are always binarized and the input activations are quantized to the desired bit width. 
For the LeNet-5 results, QNN performs marginally worse than DoReFa-Net.
DoRefa-Net achieves better stability on accuracy when quantizing down to 2 bits.
For ResNet-20 applied to CIFAR-10, quantizing from 32 bits down to 2 bits, DoReFa-Net incurs in less than 2.4\% accuracy loss, resulting in an average accuracy of 87.82\%.
QNN starts to show higher degradation on accuracy when quantizing to 2 bits.
For binarization, both methods present a significant degradation on accuracy compared to 2-bit quantization, proving once more the importance of keeping input activations at higher bit widths.

\gboxbegin{kobackteal}{koborderteal}{TAKEAWAY}{\rtasktak{kt}}
\textbf{Models can be quantized with memory savings of up to 32$\times$ with minimal accuracy degradation.}
Reducing the bit width of input activations down to 2 bits while also reducing weight bit widths results in minimal accuracy losses.
The use of fewer bits helps reduce the computational cost of performing inference in quantized neural networks.
\gboxend

\subsection{Quantization of Weights and Input Activations vs. Accuracy}
\label{sec:quant-wei-in-act}

\Cref{subfig:lenet5-qnn,subfig:lenet5-dorefa-net} show the quantization results applying QNN and DoReFa-Net, respectively, obtained for the MNIST dataset with the LeNet-5 network.
DoReFa-Net is more consistent than QNN.
The importance of keeping the input activations at a higher bit width is clearly noticeable here, especially on \Cref{subfig:lenet5-dorefa-net} that shows the results for DoReFa-Net.
Here, the blue bars at the left of the image correspond to the accuracy results obtained when the input activations were binarized.
\Cref{subfig:resnet20-qnn,subfig:resnet20-dorefa-net} show the results obtained with QNN and DoReFa-Net, respectively, when quantizing ResNet-20 applied to CIFAR-10.
Again, if the input activations are kept with bit widths greater than 1 bit, the accuracy results are consistent across all quantization plane.
The accuracy loss due to quantization of weights is minimal all the way down to 3 bits.
2- and 1-bit quantization still offer acceptable results.
This result continues to show the importance of input activations.

\gboxbegin{kobackteal}{koborderteal}{TAKEAWAY}{\rtasktak{kt}}
\textbf{The quantization of input activations has a greater effect on accuracy loss than the quantization of weights.}
\Crefrange{subfig:lenet5-qnn}{subfig:resnet20-dorefa-net} show that it is possible to reduce the bit width of weights and/or input activations and still achieve good results with a minimal loss in accuracy.
Binarization yields the worst accuracy, especially if input activations are binarized.
The quantization of input activations is more prone to reducing classification accuracy than the quantization of weights.
\gboxend

\subsection{The Importance of the First Convolutional Layer and of the Final Fully Connected Layer}

The quantization of the first convolutional layer and of the final fully connected layer is a key factor for the accuracy of the aforementioned quantization methods.
In this section, we evaluate accuracy results of quantized models on multiple levels of bit width.
Since both \ac{qnn}~\cite{qnn} and DoReFa-Net~\cite{dorefa-net} allow multiple levels of bit width quantization, we evaluate if the level of quantization applied to the parameters of these layers significantly impacts the accuracy of the network.
The networks used for this study were LeNet-5, trained with MNIST, and ResNet-20, trained with CIFAR-10.
The accuracy results obtained in this study are sumarized in Tables \ref{tab:FLL-quant-study-LeNet5} and \ref{tab:FLL-quantization-study-ResNet20}.

\begin{table}[!b]
	\centering
	\begin{tabular}{|c|cccc|}
		\hline
		\multirow{2}{*}{Method} & \multicolumn{4}{c|}{Quantization of \textunderscore{W}eights and \textunderscore{A}ctivations} \\
		& W8A8           & W4A4           & W2A2           & W1A1          \\ \hline
		\multicolumn{5}{|c|}{First and last layers NOT quantized}                                          \\ \hline
		QNN~\cite{qnn}                  & 99.61\%        & 99.56\%        & 99.49\%        & 99.33\%       \\ \hline
		DoReFa-Net~\cite{dorefa-net}    & 99.69\%        & 99.68\%        & 99.62\%        & 99.06\%       \\ \hline
		\multicolumn{5}{|c|}{First layer quantized / Last layer NOT quantized}                             \\ \hline
		QNN~\cite{qnn}                  & 99.62\%        & 99.54\%        & 99.44\%        & 99.25\%       \\ \hline
		DoReFa-Net~\cite{dorefa-net}    & 99.65\%        & 99.66\%        & 99.61\%        & 98.93\%       \\ \hline
		\multicolumn{5}{|c|}{Last layer quantized / First layer NOT quantized}                             \\ \hline
		QNN~\cite{qnn}                  & 99.59\%        & 99.57\%        & 99.38\%        & 98.98\%       \\ \hline
		DoReFa-Net~\cite{dorefa-net}    & 99.65\%        & 99.68\%        & 99.57\%        & 98.82\%       \\ \hline
		\multicolumn{5}{|c|}{First and last layers quantized}                                              \\ \hline
		QNN~\cite{qnn}                  & 99.52\%        & 99.55\%        & 99.45\%        & 98.90\%       \\ \hline
		DoReFa-Net~\cite{dorefa-net}    & 99.68\%        & 99.62\%        & 99.54\%        & 98.44\%       \\ \hline
	\end{tabular}
	\caption{Quantization of the first and last convolutional and fully connected layers. Results for LeNet-5 applied to MNIST. Single-precision results: QNN - 99.69\%; DoReFa-Net - 99.67\%}
	\label{tab:FLL-quant-study-LeNet5}
\end{table}

\begin{table}[!b]
	\centering
	\begin{tabular}{|c|cccc|}
		\hline
		\multirow{2}{*}{Method} & \multicolumn{4}{c|}{Quantization of (W)eights and (A)ctivations} \\
		& W8A8            & W4A4            & W2A2           & W1A1           \\ \hline
		\multicolumn{5}{|c|}{First and last layers NOT quantized}                                             \\ \hline
		QNN~\cite{qnn}                  & 88.41\%         & 88.90\%         & 86.37\%        & 81.35\%        \\ \hline
		DoReFa-Net~\cite{dorefa-net}    & 88.42\%         & 88.27\%         & 88.24\%        & 62.70\%        \\ \hline
		\multicolumn{5}{|c|}{First layer quantized / Last layer NOT quantized}                                \\ \hline
		QNN~\cite{qnn}                  & 88.71\%         & 88.68\%         & 85.55\%        & 79.10\%        \\ \hline
		DoReFa-Net~\cite{dorefa-net}    & 89.00\%         & 89.25\%         & 87.62\%        & 60.62\%        \\ \hline
		\multicolumn{5}{|c|}{Last layer quantized / First layer NOT quantized}                                \\ \hline
		QNN~\cite{qnn}                  & 89.35\%         & 89.04\%         & 83.64\%        & 66.47\%        \\ \hline
		DoReFa-Net~\cite{dorefa-net}    & 82.31\%         & 80.85\%         & 68.25\%        & 47.92\%        \\ \hline
		\multicolumn{5}{|c|}{First and last layers quantized}                                                 \\ \hline
		QNN~\cite{qnn}                  & 89.40\%         & 89.29\%         & 83.14\%        & 58.10\%        \\ \hline
		DoReFa-Net~\cite{dorefa-net}    & 52.62\%         & -               & -              & -              \\ \hline
	\end{tabular}
	\caption{Quantization of the first and last convolutional and fully connected layers. Results for ResNet-20 applied to CIFAR-10. Single-precision results: QNN - 91.06\%; DoReFa-Net - 90.40\%}
	\label{tab:FLL-quantization-study-ResNet20}
\end{table}

\vspace{1mm}
\noindent
\textbf{MNIST.}
Looking first at the results obtained with LeNet-5 trained with the MNIST dataset, DoReFa-Net achieves the best results when quantizing with 2 or more bits.
The effect of quantizing the first or last layers is minimal on accuracy.
When quantization is applied to both the first and last layers, there is a small degradation in accuracy.
QNN only outperforms DoReFa-Net when binarizing the model.
By default, DoReFa-Net does not quantize both layers, while QNN does.
If these layers are not quantized in the QNN, it achieves better results than DoReFa-Net when applying binarization to the model.
In conclusion, for the results obtained for LeNet-5, reducing the bit width only degrades accuracy by a small amount; the degradation resulting from quantizing both the first and last layers is less than 1\%.
The accuracy loss is never more than 2\% for all results compared to single-precision results.

\vspace{1mm}
\noindent
\textbf{CIFAR-10.}
The results obtained with ResNet-20 lead to some different conclusions.
The accuracy degradation is more evident when reducing the bit width on both methods.
When quantizing to 4 or 8 bits, DoReFa-Net can achieve good results only if the last layer is not quantized, with less than 2\% accuracy degradation, compared to single-precision.
In contrast, the accuracy degradation is significant if the last layer is quantized, ranging between 8 and 10\%.
In the specific case of 2-bit quantization applied to both layers, DoReFa-Net does not converge to an optimal solution.
QNN, in the same case of 4- or 8-bit quantization, achieves good results, independent of the quantization of the first and/or last layers, with less than 3\% accuracy degradation.
For 2-bit quantization, QNN continues to achieve good results, if at least the last layer is not quantized.
If quantization is only applied on the last layer, the accuracy degradation is significant.
If quantizing both layers, DoReFa-Net does not converge to an optimal solution.
In the case of QNN, it is more stable when quantizing these layers, but if these are not quantized, especially the last layer, the accuracy results are better, as mentioned above.
QNN, for binarization, continues to show 
better results
than DoReFa-Net.
DoReFa-Net shows difficulties achieving good results, even when both layers are not quantized.
The importance of at least the last layer not being quantized continues to be evident.
QNN achieves good results when, at least, the last layer is not quantized.
If both layers are not quantized, QNN achieves 81.35\% accuracy compared to 91.06\% in single-precision, less than 10\% accuracy degradation.
In conclusion, for the ResNet-20 results, QNN continues to show its strength, achieving more stable results and good results for binarized models.
The last layer is important, with both methods achieving better results when this layer is not quantized.
DoReFa-Net shows difficulties when the last layer is quantized.
If both layers are quantized, DoReFa-Net might not be able to converge on an optimal solution.

\gboxbegin{kobackteal}{koborderteal}{TAKEAWAY}{\rtasktak{kt}}
\textbf{Different networks tolerate the quantization of first and last layers differently.}
In relatively small networks (e.g., LeNet-5), the effect of quantizing the first and last layers is minimal.
In contrast, the accuracy of larger networks can be severely impacted by quantizing the last layer, and it is therefore important to preserve full precision in this case.
\gboxend

\section{Making the Leap to Hardware}

The results reported above demonstrate that the reduction of bit width allows 1) the reduction of model size and 2) the reduction of logic operations performed.
One can, therefore, exploit the implementation of specialized hardware accelerators for inferencing in \acp{cnn}. These implementations can reduce the requirements both in operations-related hardware and memory. Ultimately, lower energy consumption is achieved.

Despite \acsp{cpu} offering lower capabilities in terms of number of cores, they still offer great performance due to their high frequency clock cycle. Some works explore quantization to improve \acs{cpu} performance~\cite{cpu_accelerator1, bitstream, cpu_accelerator2, cpu_accelerator3, dabnn, dsq}. It is also possible to develop tools that are hardware agnostic and can accelerate deep learning tasks on both \acsp{cpu} and \acp{gpu} \cite{bmxnet}.

Most of the \acp{cnn} are trained on \acp{gpu}. \acp{gpu}, can also be exploited to take advantage of \acp{qcnn}.
NVIDIA introduced TensorRT, a framework capable of using tensor cores to perform operations with 2D tensors \cite{tensor-core}.
Tensor cores perform \acp{gemm} faster than it was previously possible, improving throughput and/or reducing latency.
Tensor cores also work with mixed precision, i.e., lower than single precision operations, improving throughput and latency further more.
\acp{qcnn} can exploit this new technology. Several works already explore mixed precision to improve throughput, latency and energy to performance ratio \cite{mixed_prec_training, mixed_prec_training2, mixed_prec_solvers, bfloat16, ampt_ga, mp_solver_matrix}. Some works even explored mixed precision on massive \ac{gpu} clusters for \acl{hpc} achieving significant gains in throughput per energy consumption \cite{mp_cluster, mp_cluster2, mp_cluster3}.
Despite recent developments, \acp{gpu} still have a fixed amount of hardware built into them, which cannot not always can be fully utilized.
This leads to lower efficiency.

\subsection{Towards Hardware Specialization}

In contrast to CPUs and \acp{gpu}, the flexible and reconfigurable nature of \acp{fpga} allows the prototyping and implementation of specific solutions for each project/task, reducing the required hardware.
Depending on the level of quantization, \acp{fpga} allow the tuning of the level of optimization towards more efficient solutions, as reduced bit widths mitigate hardware design complexity.
In some cases, such as binarization, most operations occurring in \acp{cnn} can be implemented efficiently using lookup tables.
\acp{fpga} reduce the energy consumption and improve latency and throughput, compared to \acp{gpu}.
Most works focus on \ac{fpga} accelerator implementations for \acp{bnn} \cite{semi_bin_bwn_fpga, fp_bnn_fpga, BNN_FPGA_implementation1, finn_fpga, finn_fpga2, fast_bnn_fpga, rebnet_fpga, binaryeye_fpga, fbna_fpga, dlbnn_fpga, memory_b_bnn_fpga, bnn_no_bn_fpga} but some also give their attention to other quantization bit widths \cite{low_precision_fpga, finn_r_fpga, ternary_fpga, lightnn_fpga}.

\acp{asic}, like \acp{fpga}, allow the implementation of task-specific hardware acceleration \cite{yodann-asic-accelerator, cnn-fpga-asic-accelerator, deepopt-asic, dnn_accelerator, eyeriss, eyeriss2, eyeriss_v2}.
Although they are not reprogrammable such as \acp{fpga}, they can offer better efficiency and throughput, compared to \acp{fpga}.
Low power \acp{soc} with good deep learning accelerators allow more efficient edge computation \cite{asic_soc}.

\subsection{Addressing the Data Movement Bottleneck}

The reduction in model size afforded by \acp{qnn} considerably mitigates the overall requirements of data movement between the system's compute and memory units, which is a well-known major contributor to energy and performance overheads in modern systems~\cite{boroumand2018google}.
However, this bottleneck can be addressed more fundamentally by processing data closer to where it resides; this is the guiding principle behind processing-in-memory (PIM) architectures, an emerging computational paradigm that confers substantial throughput and energy benefits.
Under the \ac{pim} computing paradigm, data is processed either near-memory (Processing Near Memory, PNM), or using-memory (Processing Using Memory, PUM).
Under the Processing Near Memory (PNM) paradigm, computation takes place close to where data resides, but in a different medium~\cite{devaux2019true,vcache,kwon202125,jun2017hbm,pawlowski2011hybrid}. This medium implements dedicated compute units, which can either be 1) general-purpose, in which case they are usually implemented as simple in-order cores, or 2) special-purpose, in which case performance and energy efficiency are maximized at the expense of general applicability.
In contrast, Processing Using Memory (PUM) is a paradigm that takes advantage of the physical properties of the storage medium (i.e., the memory cells) to perform computation.
Its range of applicability therefore tends to be narrower, but PUM mechanisms typically provide superior throughput and energy efficiency.
Many prior PUM works~\cite{rowclone,bitwise-operations,pim-instructions,drisa,ambit,neural-cache,dima,parapim,rebnn,rtl_lib-compiler,mp_in_memory_computation,float_pim,ferreira2021pluto} introduce mechanisms which can be readily applied to perform inference in \acp{bnn}, since arithmetic operations in these networks can be expressed using bitwise logic operations, rather than conventional integer arithmetic.

\section{Conclusion}

We introduce \mechanism, a framework that enables the user-friendly, systematic evaluation of a wide gamut of \acp{qcnn} by streamlining their training and testing procedures.
Using \mechanism, we implement and evaluate a total of $4500$ unique network configurations, using various methods, quantization levels, and datasets.
Our results show that \acp{qcnn} achieve near state-of-the-art accuracy while simultaneously reducing the memory and computational footprint of modern \ac{cnn} models by up to $32\times$.

All source code and results have been open-sourced.
We actively encourage the community to build upon the $20000$+ hours of active compute time we provide as a starting point, and to contribute towards the development of additional models, quantization methods, and datasets, as well as their evaluation.
Through a community-driven effort it will be possible to leverage \mechanism to create the world's largest repository of \ac{qcnn} configurations, which will aid the prototyping, development, and application of \acp{qcnn}.

\vspace{3mm}
\begin{tcolorbox}[colback=repoboxback,colframe=repoboxborder,title=\textbf{\emph{Getting Started with \mechanism}}]
All source code and results are available at \textbf{\href{https://github.com/IT-Coimbra/RedBit}{https://github.com/IT-Coimbra/RedBit}}.

Developers interested in contributing should refer to this repository, which includes instructions for reproducing our results and for contributing additional quantization models, methods, and results.
\end{tcolorbox}

\ifCLASSOPTIONcompsoc
  \section*{Acknowledgments}
\else
  \section*{Acknowledgment}
\fi

The authors would like to thank the Laboratory for Advanced Computing of University of Coimbra \linebreak (\url{https://www.uc.pt/lca}) for the HPC and consulting resources provided, which made the attainment of the results reported in this paper possible.
This work was also supported by Instituto de Telecomunicações and Fundação para a Ciência e a Tecnologia, Portugal, under grants UIDB/50008/2020 and EXPL/EEI-HAC/1511/2021.

\ifCLASSOPTIONcaptionsoff
  \newpage
\fi

\bibliographystyle{IEEEtran}
\bibliography{main}

\vspace{-10mm}
\begin{IEEEbiography}[{\includegraphics[width=1in,height=1.25in,clip,keepaspectratio]{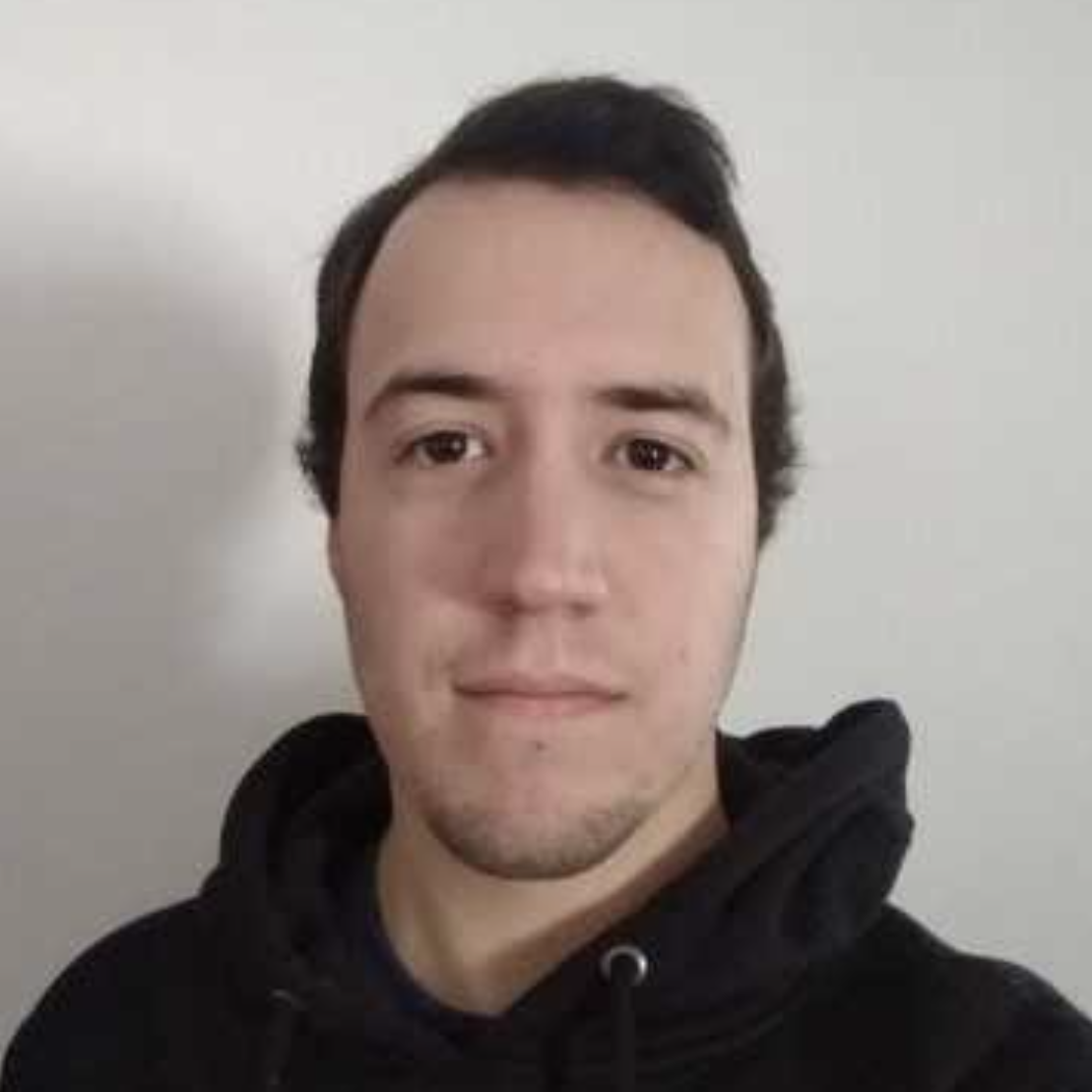}}]{Andre Santos}
Andre Santos is pursuing an Integrated Master’s Degree in Electrical and Computer Engineering at the University of Coimbra, Portugal, with a thesis that addresses the impact of quantization on AI hardware and classification accuracy. His research interests include quantization and machine learning architectures and systems.
\end{IEEEbiography}

\vspace{-10mm}
\begin{IEEEbiography}[{\includegraphics[width=1in,height=1.25in,clip,keepaspectratio]{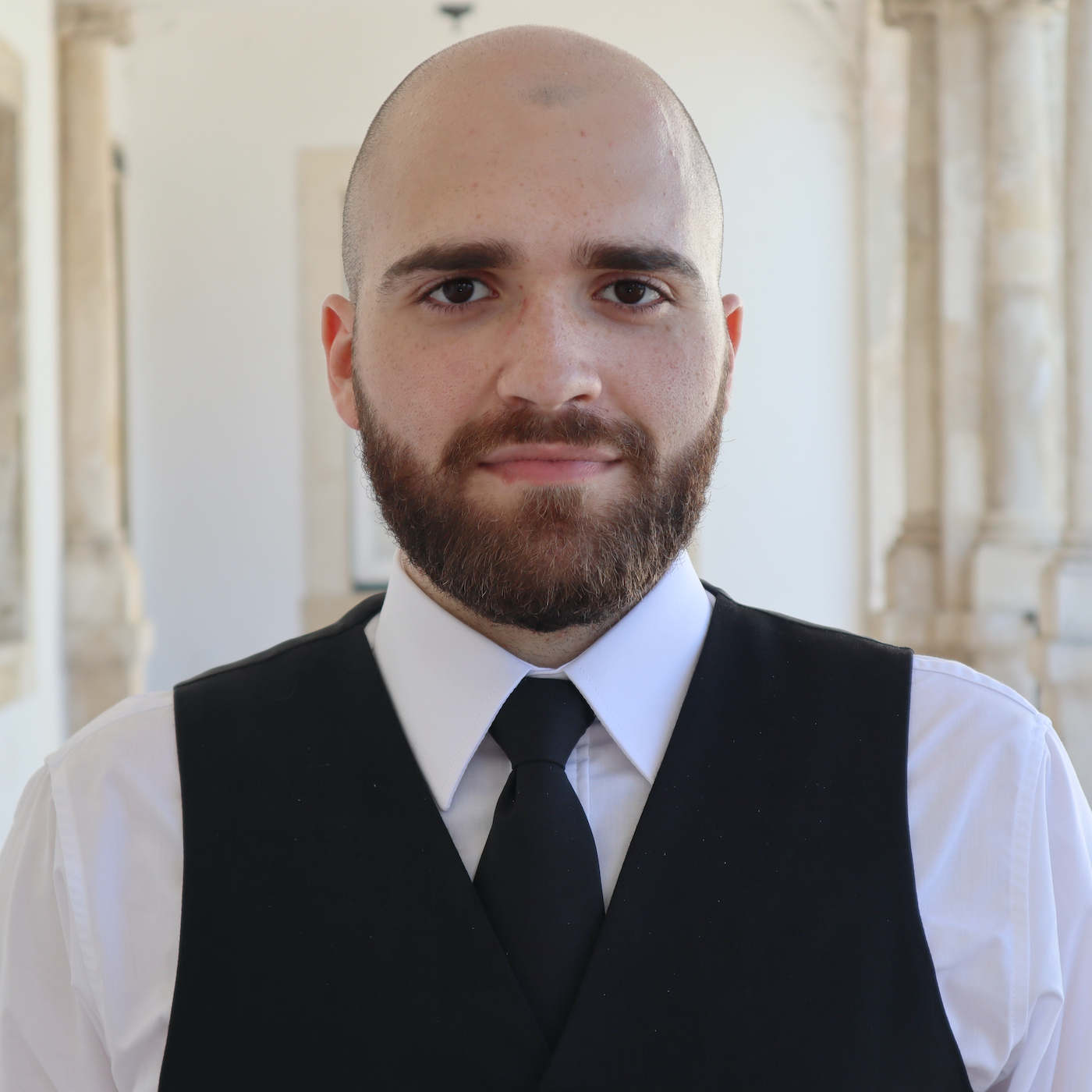}}]{João Dinis Ferreira}
João Dinis Ferreira holds a Bachelor’s Degree in Electrical and Computer Engineering from the University of Coimbra, where he graduated in the top 3\% of his class. He is currently a research and teaching assistant at ETH Zurich, where he is pursuing a Master’s Degree in Electrical Engineering and Information Technology. His research interests include memory systems, processing-in-memory and machine learning hardware.
\end{IEEEbiography}

\vspace{-10mm}
\begin{IEEEbiography}[{\includegraphics[width=1in,height=1.25in,clip,keepaspectratio]{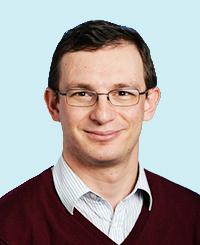}}]{Onur Mutlu}
Onur Mutlu is a Professor of Computer Science at ETH Zurich. His current broader research interests are in computer architecture, systems, hardware security, and bioinformatics. He obtained his PhD and MS in ECE from the University of Texas at Austin and BS degrees in Computer Engineering and Psychology from the University of Michigan, Ann Arbor. He started the Computer Architecture Group at Microsoft Research (2006-2009), and held various product and research positions at Intel Corporation, Advanced Micro Devices, VMware, and Google. He received the IEEE Computer Society Edward J. McCluskey Technical Achievement Award, the ACM SIGARCH Maurice Wilkes Award, the inaugural IEEE Computer Society Young Computer Architect Award, the inaugural Intel Early Career Faculty Award, US National Science Foundation CAREER Award, Carnegie Mellon University Ladd Research Award, faculty partnership awards from various companies, and a healthy number of best paper or "Top Pick" paper recognitions at various computer systems, architecture, and hardware security venues.
\end{IEEEbiography}

\begin{IEEEbiography}[{\includegraphics[width=1in,height=1.25in,clip,keepaspectratio]{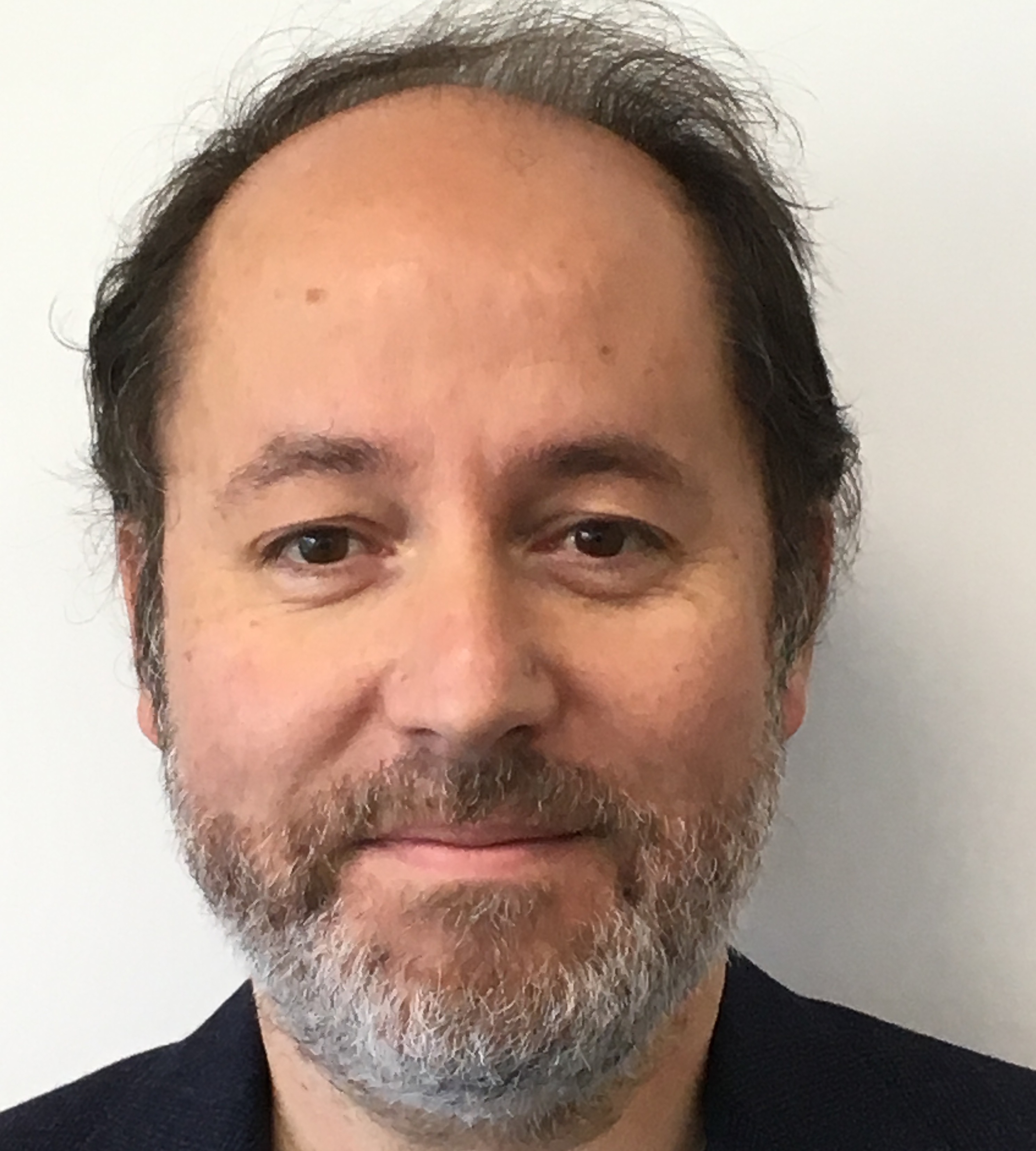}}]{Gabriel Falcao} (S'07--M'10--SM'14) received the Ph.D. degree from the University of Coimbra, in 2010, where he is currently a Tenured Assistant Professor with the Department of Electrical and Computer Engineering. In 2011 and 2017, he was a Visiting Professor with EPFL, and in 2018, he was a Visiting Academic with ETHZ, both in Switzerland. He is also a Researcher with Instituto de Telecomunica\c{c}\~{o}es. His research interests include parallel computer architectures, energy-efficient processing, GPU- and FPGA-based accelerators, and compute-intensive signal processing applications including machine learning. In 2020 Gabriel Falcao was General Co-Chair of the IEEE SiPS and in 2021 Local Chair of Euro-Par 2021. He is a Senior Member of the IEEE, Member of the IEEE Signal Processing Society and a Full Member of the HiPEAC Network of Excellence.
\end{IEEEbiography}

\balance

\end{document}